\documentclass[10pt,twocolumn,letterpaper]{article}

\usepackage{wacv}
\usepackage{times}
\usepackage{epsfig}
\usepackage{graphicx}
\usepackage{amsmath}
\usepackage{amssymb}
\usepackage{multirow}

\usepackage{caption}
\usepackage{subcaption}
\usepackage{color}
\usepackage{tikz}
\usetikzlibrary{arrows}
\usetikzlibrary{calc}
\usepackage{relsize}
\usepackage[accsupp]{axessibility}
\usepackage{authblk}
 
\newcommand\blfootnote[1]{%
  \begingroup
  \renewcommand\thefootnote{}\footnote{#1}%
  \addtocounter{footnote}{-1}%
  \endgroup
}

\def\wacvPaperID{1050} 
\wacvfinalcopy 

\ifwacvfinal
\def\assignedStartPage{1} 
\fi

\ifwacvfinal
\usepackage[breaklinks=true,bookmarks=false]{hyperref}
\else
\usepackage[pagebackref=true,breaklinks=true,colorlinks,bookmarks=false]{hyperref}
\fi

\ifwacvfinal
\else
\fi

\begin{document}

\title{CeyMo: See More on Roads - A Novel Benchmark Dataset for Road Marking Detection}
\author{Oshada Jayasinghe}
\author{Sahan Hemachandra}
\author{Damith Anhettigama}
\author{Shenali Kariyawasam}
\author{\space\space \space \space  \space  \space     \space    \space  \space \space\space \space \space  \space  \space     \space    \space  \space Ranga Rodrigo}
\author{Peshala Jayasekara}
\affil{Department of Electronic and Telecommunication Engineering, \\ University of Moratuwa, Sri Lanka \\ \vspace{1.5mm} \tt\small oshadajayasinghe@gmail.com, sahanhemachandra@gmail.com, damithkawshan@gmail.com, shenali1997@gmail.com, ranga@uom.lk, peshala@uom.lk}
\maketitle

\thispagestyle{plain}
\pagestyle{plain}

\begin{abstract}
    In this paper, we introduce a novel road marking benchmark dataset for road marking detection, addressing the limitations in the existing publicly available datasets such as lack of challenging scenarios, prominence given to lane markings, unavailability of an evaluation script, lack of annotation formats and lower resolutions. Our dataset consists of 2887 total images with 4706 road marking instances belonging to 11 classes. The images have a high resolution of $1920 \times 1080$ and capture a wide range of traffic, lighting and weather conditions. We provide road marking annotations in polygons, bounding boxes and pixel-level segmentation masks to facilitate a diverse range of road marking detection algorithms. The evaluation metrics and the evaluation script we provide, will further promote direct comparison of novel approaches for road marking detection with existing methods. Furthermore, we evaluate the effectiveness of using both instance segmentation and object detection based approaches for the road marking detection task. Speed and accuracy scores for two instance segmentation models and two object detector models are provided as a performance baseline for our benchmark dataset. The dataset and the evaluation script will be publicly available.
\end{abstract}
\vspace{-1em}
\blfootnote{\textcopyright \ 2022 IEEE. Personal use of this material is permitted.
  Permission from IEEE must be obtained for all other uses, in any current or future 
  media, including reprinting/republishing this material for advertising or promotional 
  purposes, creating new collective works, for resale or redistribution to servers or 
  lists, or reuse of any copyrighted component of this work in other works.}

\vspace{-1em}

\section{Introduction}

\begin{figure}
     \centering
     \begin{subfigure}[b]{0.32\linewidth}
         \centering
         \includegraphics[width=.98\linewidth]{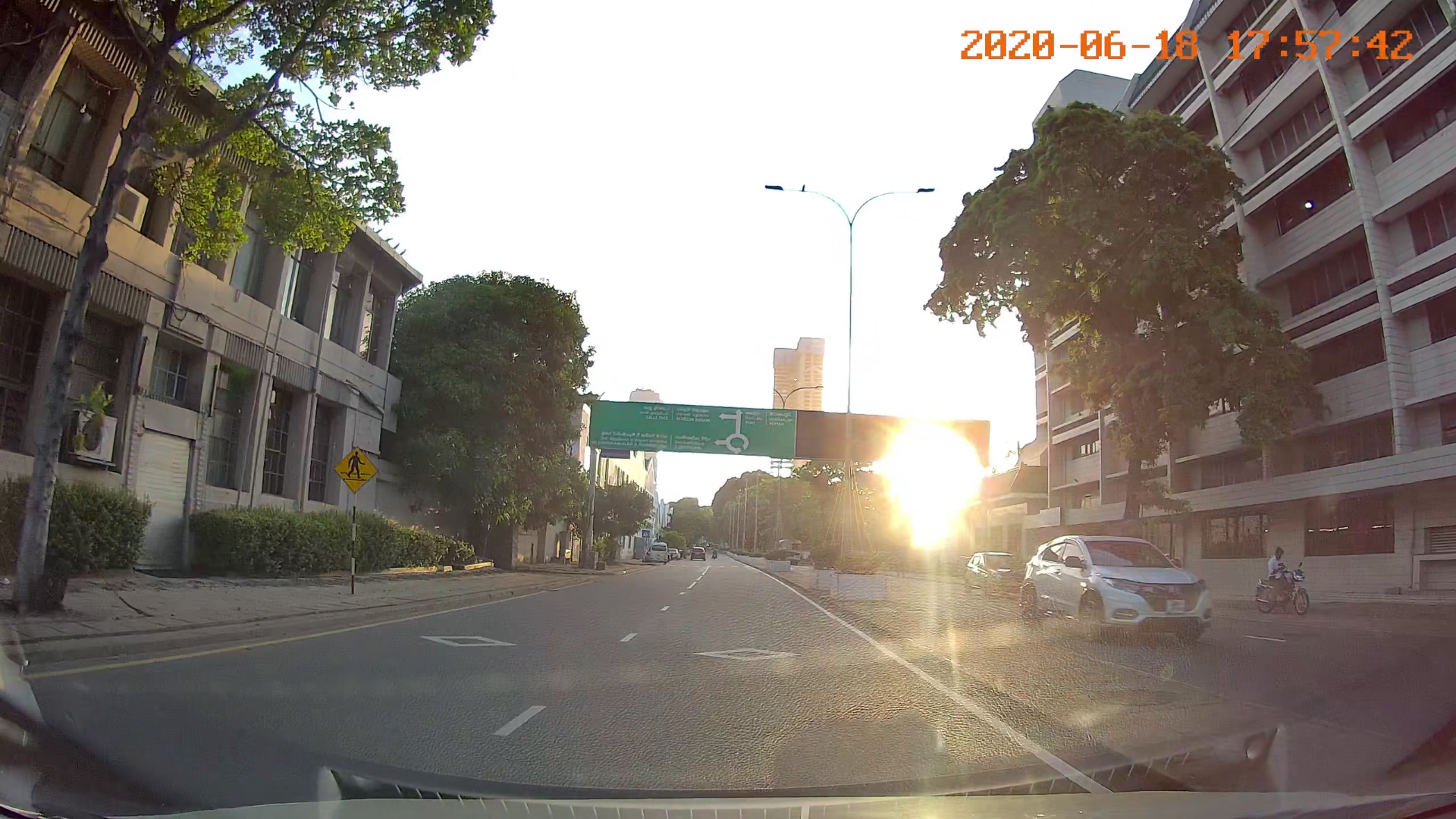}
         \caption{}
     \end{subfigure}%
     \begin{subfigure}[b]{0.32\linewidth}
         \centering
         \includegraphics[width=.98\linewidth]{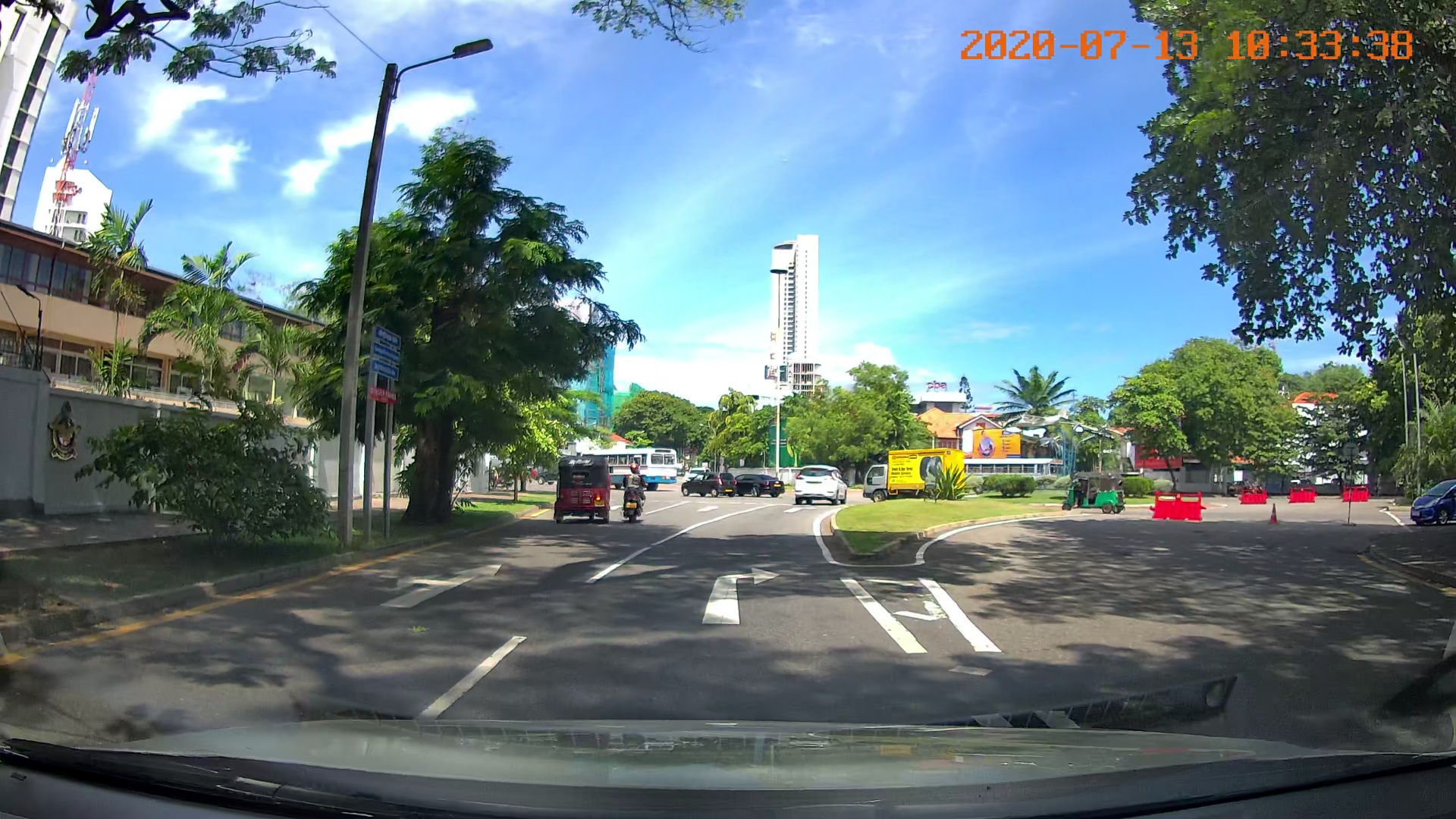}
         \caption{}
     \end{subfigure}%
     \begin{subfigure}[b]{0.32\linewidth}
         \centering
         \includegraphics[width=.98\linewidth]{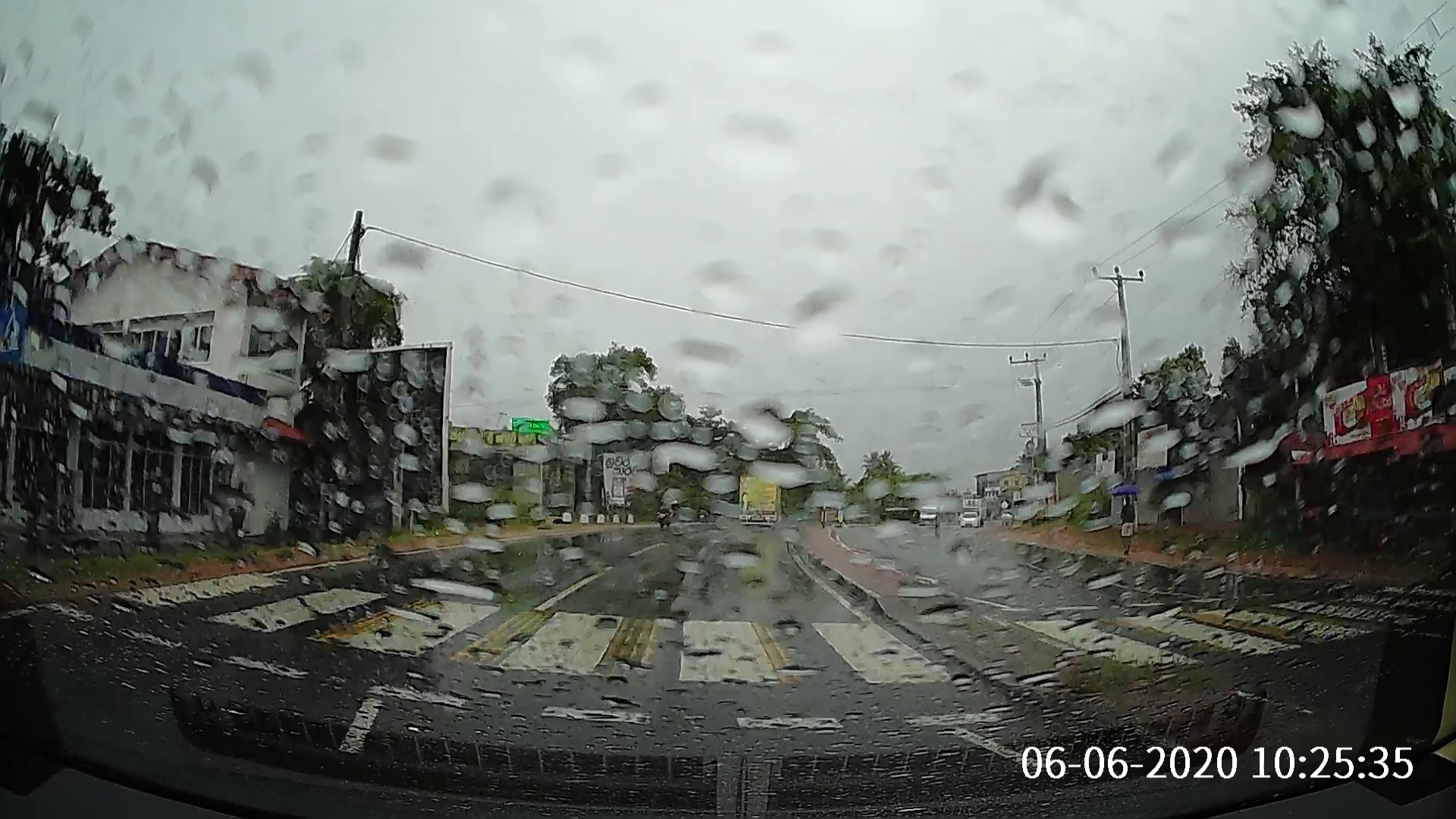}
         \caption{}
     \end{subfigure}%
     
     \begin{subfigure}[b]{0.32\linewidth}
         \centering
         \includegraphics[width=.98\linewidth]{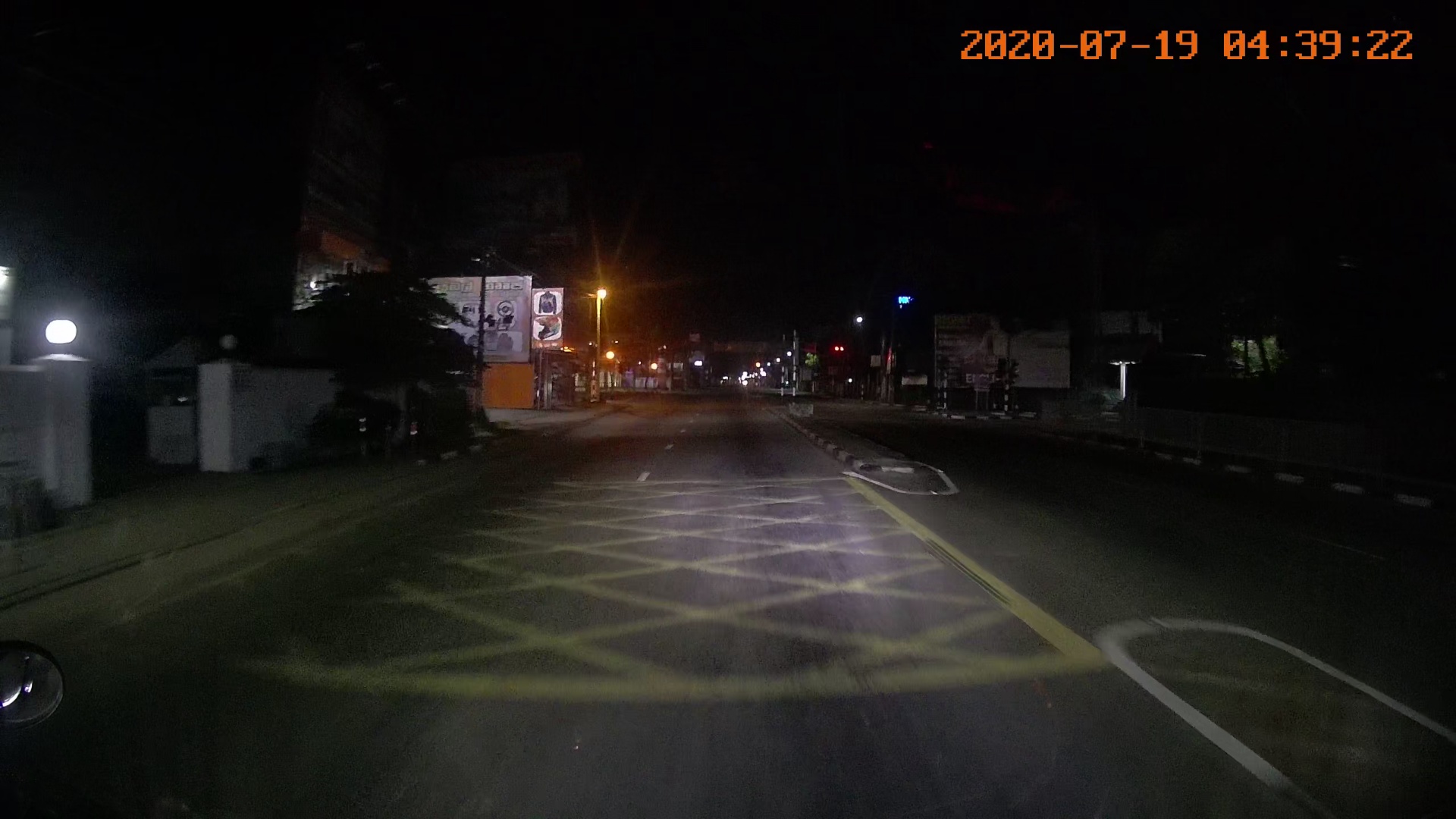}
         \caption{}
     \end{subfigure}%
     \begin{subfigure}[b]{0.32\linewidth}
         \centering
         \includegraphics[width=.98\linewidth]{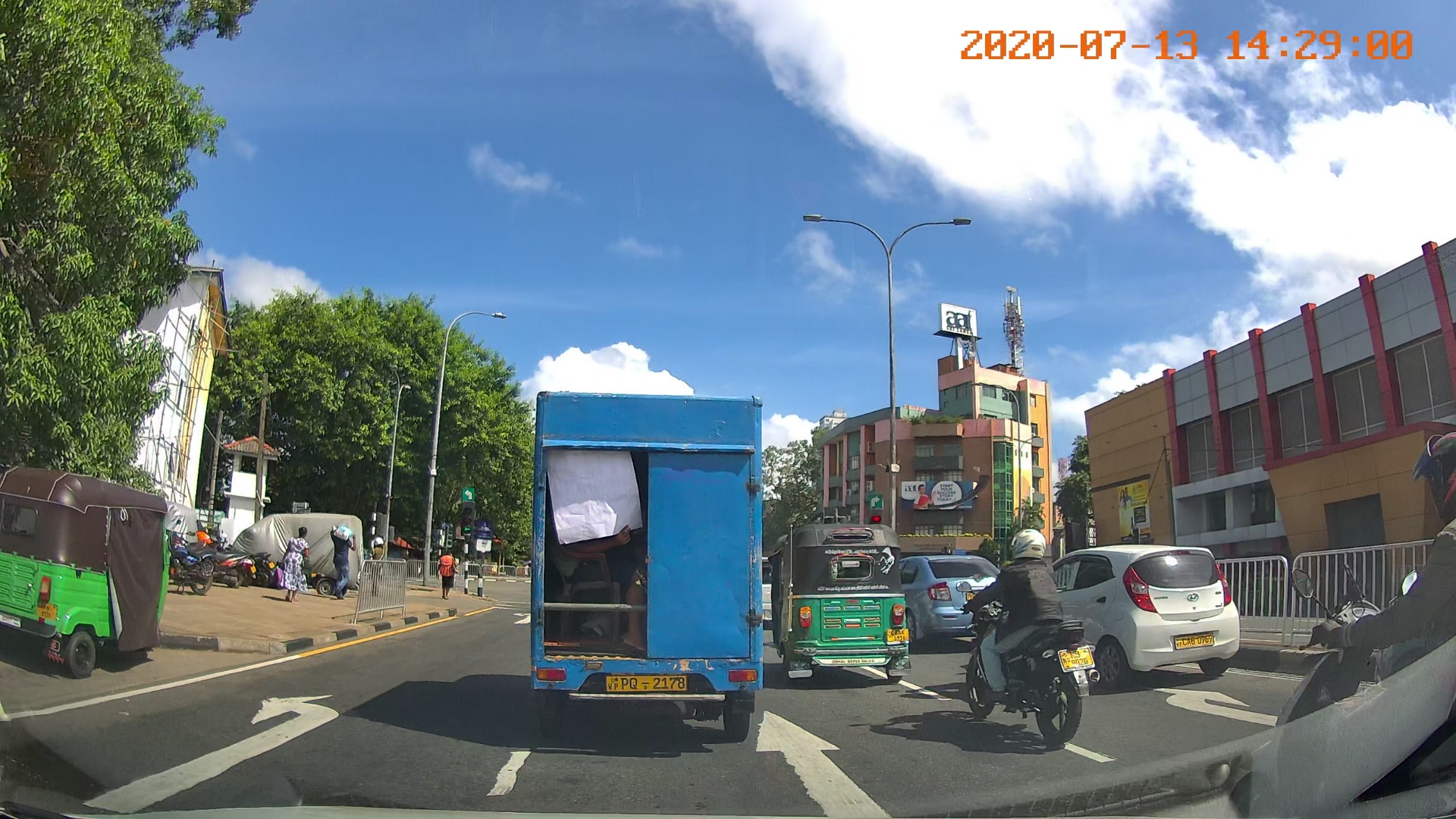}
         \caption{}
     \end{subfigure}%
     \begin{subfigure}[b]{0.32\linewidth}
         \centering
         \includegraphics[width=.98\linewidth]{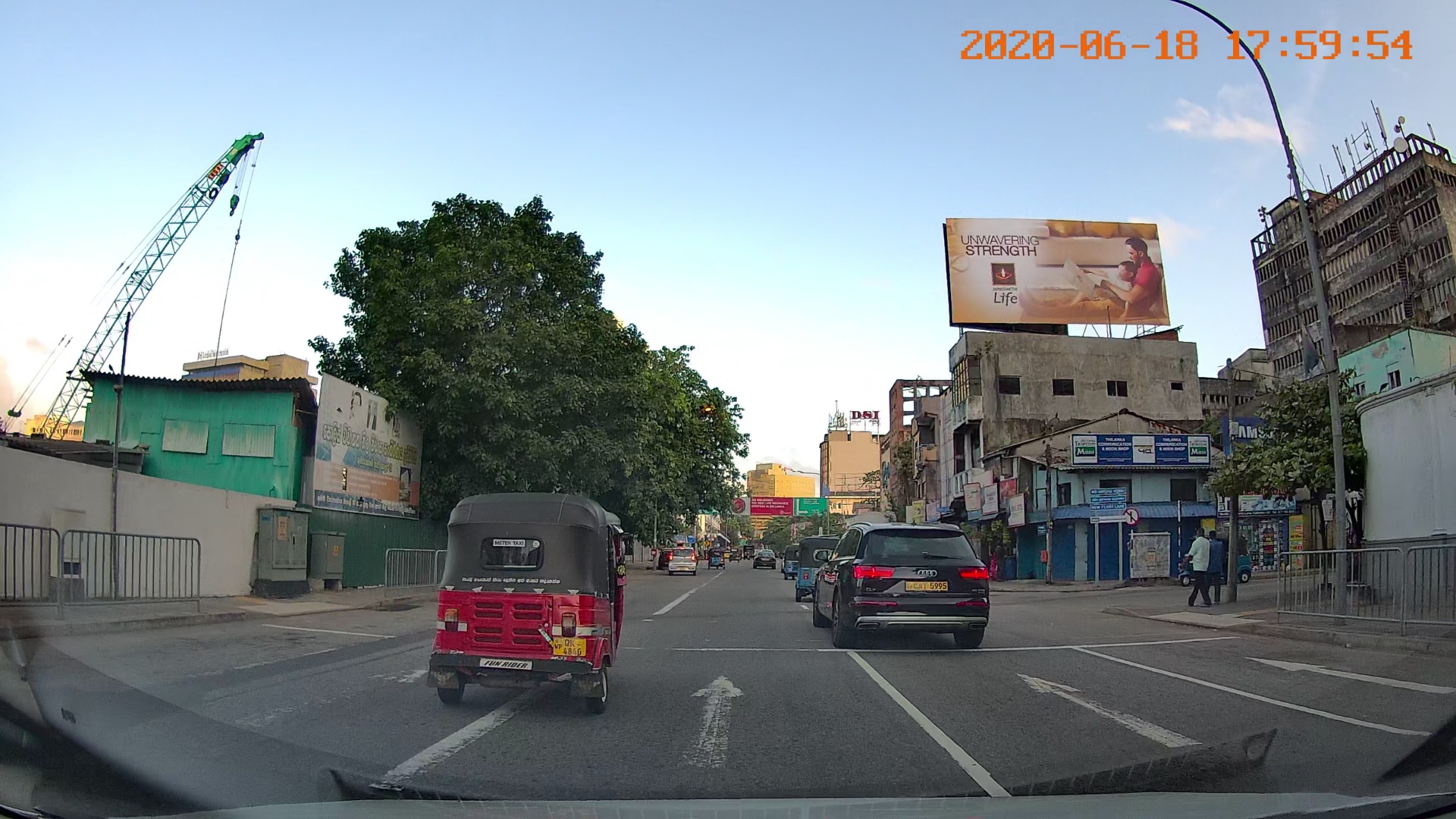}
         \caption{}
     \end{subfigure}%
    \caption{
    Challenging scenarios present in our dataset. (a) Dazzle light (b) Shadow (c) Rain (d) Night (e) Occlusion (f) Deteriorated road markings}
    
    \label{fi:challenging_scenarios}
\vspace{-1em}
\end{figure}

\begin{table*}[t]
\begin{center}
\begin{tabular}{|l|c|c|c|c|c|c|}
\hline
\textbf{Dataset} & \textbf{Year} & \textbf{Images} & \textbf{Classes} & \textbf{Location} & \textbf{Annotation Format} \\
\hline\hline
Road Marking \cite{ananth} & 2012 & 1443 & 11 & USA & Bounding Box annotations (TXT) \\
\hline
TRoM \cite{TROM} & 2017 & 712 & 19 & China & Pixel-level annotations (PNG) \\
\hline
VPGNet \cite{VPGNet} & 2017 & 21097 & 17 & Korea & Pixel-level and Grid-level annotations (MAT)  \\
\hline
\end{tabular}
\vspace{-0em}
\caption{Summary of existing road marking detection datasets.}
\label{Tab:summary}
\end{center}
\vspace{-1em}
\end{table*}

Understanding traffic regulations imposed by traffic symbols such as traffic signs, traffic lights, lane and road markings can be considered as a fundamental perception task involved in the development of advanced driver assistance systems (ADAS) and autonomous vehicles. Road markings refer to the symbols and text painted on the road surface, which assist the drivers to safely navigate on roads by regulating the traffic. Developing robust road marking detection algorithms is a challenging task due to occlusions, illumination changes, shadows, varying weather conditions and deterioration of road signs with time.

Though the detection and recognition of road markings is a vital task, it is often a less researched area, mainly due to the lack of publicly available datasets and limitations present in the existing datasets. The dataset introduced by \cite{ananth} is used in most of the earlier work done on road marking detection \cite{chen2015road,Kheyrollahi,IPMbased12,Srikanthan}. Though their dataset consists of 1443 images, its diversity is limited since many adjacent frames covering the same scenario have been annotated. Undefined train-test split and unavailability of an evaluation script can be identified as further issues. Some of the recent deep learning based methods \cite{VPGNet,TROM} tackle the lane detection and road marking detection as a single segmentation task. Therefore, the datasets such as VPGNet \cite{VPGNet} and TRoM \cite{TROM} contain annotations for both lane and road markings together. However, more focus is given to the lane detection task and the frequency of instances for road marking classes are much less than that for lanes. Moreover, the limited annotation formats available and the lack of proper evaluation metrics have made it difficult to accomplish novel developments and to compare with existing road marking detection approaches.

Having identified the requisite for a common benchmark for road marking detection, we introduce the CeyMo road marking dataset consisting of 2887 images and 4706 instances belonging to 11 road marking classes. As illustrated in Figure \ref{fi:challenging_scenarios}, the dataset covers a wide variety of challenging urban, sub-urban and rural road scenarios, and the test set is divided into six categories: normal, crowded, dazzle light, night, rain and shadow. The image annotations are provided as polygons, bounding boxes and segmentation masks, such that it will encourage a broad range of research in the road marking detection domain. Furthermore, we provide two evaluation metrics along with an evaluation script for the dataset, facilitating direct comparison of diverse road marking detection approaches.


Most of the existing work done on road marking detection \cite{wacv,Greenhalgh,chen2015road,Kheyrollahi,ding2020comprehensive,Suhr2015FastSR} generate candidate regions first, and then recognize the regions using machine learning based algorithms. End-to-end deep learning based instance segmentation and semantic segmentation networks have been used in recent works, \cite{TROM} and \cite{VPGNet}. The use of end-to-end object detector models to detect road markings is faster and more efficient, yet a less researched approach. We investigate the effectiveness of both instance segmentation and object detection based approaches for detecting road markings in our dataset. We use two Mask R-CNN \cite{mask_rcnn} based network architectures under the instance segmentation based approach and two SSD \cite{SSD} based object detector models, along with inverse perspective transform (IPT) are used under the object detection based approach. The inference speeds and the class-wise, scenario-wise and overall accuracy values of the four models are provided as a performance baseline. In summary, the contributions of this paper are as follows:
\begin{itemize}
  \item We introduce the CeyMo road marking dataset covering a wide variety of challenging scenarios and addressing the limitations present in existing publicly available datasets. The dataset is provided with three annotation formats, and an evaluation script to facilitate subsequent research on road marking detection. 
  
  \item We evaluate the approaches of utilizing both instance segmentation and object detection based network architectures for the road marking detection task and provide results in terms of speed and accuracy for a set of selected models on our benchmark dataset.
\end{itemize}

The rest of the paper is organized as follows: Section \ref{sec:related_works} presents related work. In Section \ref{sec:dataset}, we provide details of our benchmark dataset, while the proposed detection pipelines and employed methods are discussed in Section \ref{sec:method}. The experimental details and results are presented in Sections \ref{sec:Experiments} and \ref{sec:results}, while Section \ref{sec:conclusion} draws important conclusions.

\section{Related Work}
\label{sec:related_works}

In this section, we analyze the existing publicly available road marking detection datasets and different algorithms and implementations carried out for the road marking detection task. 

\subsection{Datasets}
\label{ssec:Datasets}

\begin{figure*}
     \centering
     \begin{subfigure}[b]{0.25\linewidth}
         \centering
         \includegraphics[width=.98\linewidth]{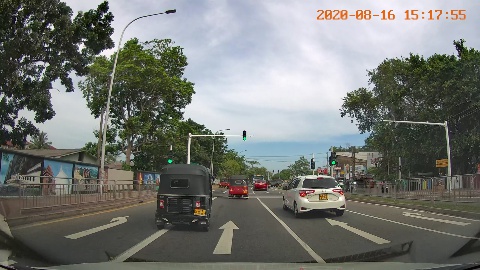}
         \caption{Image (JPG)}
     \end{subfigure}%
     \begin{subfigure}[b]{0.25\linewidth}
         \centering
         \includegraphics[width=.98\linewidth]{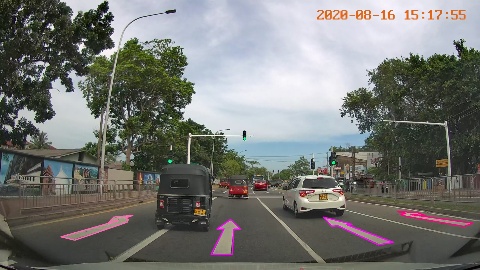}
         \caption{Polygon (JSON)}
     \end{subfigure}%
     \begin{subfigure}[b]{0.25\linewidth}
         \centering
         \includegraphics[width=.98\linewidth]{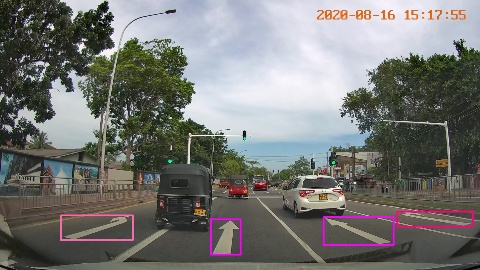}
         \caption{Bounding Box (XML)}
     \end{subfigure}%
     \begin{subfigure}[b]{0.25\linewidth}
         \centering
         \includegraphics[width=.98\linewidth]{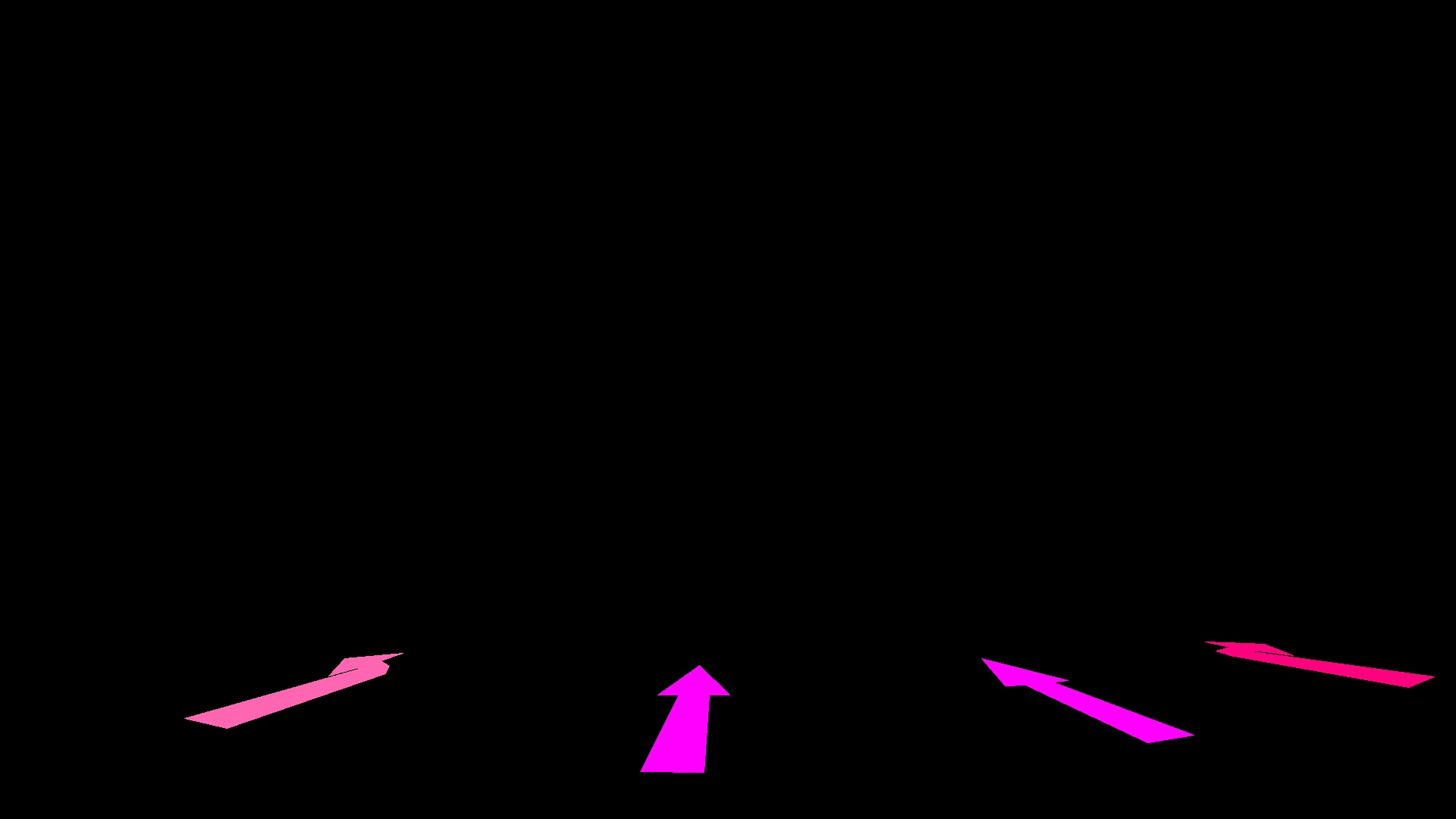}
         \caption{Segmentation Mask (PNG)}
     \end{subfigure}%
    \caption{ 
    Annotation formats provided with our dataset.}
    
    \label{fi:annotations}
\vspace{-0.5em}
\end{figure*}

As the first publicly available dataset for road marking detection, \cite{ananth} introduces a road marking dataset that contains 1443 annotated images covering 11 different road marking classes. The images have a relatively low resolution of $800 \times 600$ and the bounding box annotations for all road marking instances in all images are provided in a single text file. As their main focus relies on image processing based approaches, they do not provide separate train and test sets, and a clear evaluation metric is not specified.

Tsinghua Road Marking Dataset (TRoM) \cite{TROM} contains 712 images covering 19 different lane and road marking classes. Pixel-level semantic segmentation based annotations are provided in the PNG format. The dataset consists of 512 images in the train set, 100 images in the validation set and 100 images in the test set. Less number of images and road marking instances in this dataset might be insufficient for recent deep learning based network architectures. Furthermore, the annotation format and the evaluation metric of this dataset are designed for semantic segmentation based approaches, which limits its usability for diverse non-lane road marking detection algorithms.

VPGNet \cite{VPGNet} is a large dataset which consists of about 20000 images covering 17 lane and road marking classes. Pixel-level and grid-level annotations of the lanes and road markings including the vanishing point of the lanes are presented in the MAT file format. As the evaluation metric, the intersection over union (IoU) of the predicted cells with the ground truth grid cells are considered. However, only 8 road marking classes are covered and 52975 instances are annotated as a single class named ``other road markings". Their main focus lies on vanishing point guided lane detection, and this dataset is not widely used for road marking detection in subsequent research.    

Most of the other implementations \cite{Greenhalgh,Suhr2015FastSR,Kheyrollahi,ding2020comprehensive} use their own datasets, which are not publicly available. A summary of the three main publicly available road marking detection datasets are presented in Table \ref{Tab:summary}.

\subsection{Algorithms}

Most of the work done on road marking detection rely on classical image processing techniques combined with simple machine learning algorithms. The usual detection pipeline includes image pre-processing, regions of interest (ROI) generation, feature extraction, and classification using machine learning algorithms. A review of non-lane road marking detection and recognition algorithms is present in \cite{review}. Rectifying the original image using inverse perspective transform (IPT) \cite{ananth,Greenhalgh,IPMbased12,Kheyrollahi} is a commonly used pre-processing technique. As an alternative to IPT, \cite{Suhr2015FastSR} suggests that the search area can be reduced using the lane information. However, this may result in poor performance since road marking detection accuracy directly depends on lane detection accuracy.

Maximally stable extremal regions (MSER) \cite{MSER} are used as possible candidate regions in \cite{ananth}, and histogram of oriented gradients (HOG) feature descriptors are used to build a template pool for each class. At inference time, each image is compared with all template images to assign the classes. However, supervised learning methods usually perform better than template matching methods, especially in complex scenarios. MSER \cite{MSER} regions and HOG features have also been used with a support vector machine (SVM) classifier in \cite{Greenhalgh}, to recognize symbol based road markings. A separate optical character recognition (OCR) algorithm is used to recognize text based road markings. However, having different approaches for different road markings may result in a computational redundancy. Both of these methods include HOG feature extraction, which is a time consuming process.

Binarized normed gradients (BING) \cite{cheng2014bing} objectiveness estimation algorithm is used to generate possible road marking region proposals in \cite{chen2015road}. They use PCANet \cite{chan2015pcanet} and SVM integrated classifier to recognize the road markings. The main drawback of this method is the lower localization accuracy since BING \cite{cheng2014bing} usually results in larger proposal regions.  Logistic regression has been used with PCANet \cite{chan2015pcanet} in \cite{wacv}, to improve the classification accuracy. A shallow convolutional neural network (CNN) is also introduced as an alternative classifier for road marking recognition. After identifying MSER \cite{MSER} regions, a density based clustering algorithm is used to merge them to obtain road marking proposal regions for the classifier. However, they use many pre-processing techniques to obtain region proposals and the PCANet \cite{chan2015pcanet} or the shallow CNN classifier is only used for the recognition part.

End-to-end deep learning based networks have not been widely used in the domain of road marking detection. A convolutional neural network model which combines ResNet-101 \cite{resnet} and a pyramid pooling ensemble, is used in \cite{TROM}, to obtain lanes and road markings as semantic segmentation outputs. Their architecture achieves average results on the TRoM \cite{TROM} dataset which can be considered as a performance baseline. VPGNet \cite{VPGNet} is a CNN based architecture for detecting lanes and road markings simultaneously. They address the road marking detection as a grid regression task followed by grid sampling and box clustering as post-processing techniques for merging grid cells. However, their focus is more on lane detection and vanishing point prediction tasks, and quantitative results are only provided for four road marking classes.



\section{Benchmark Dataset}
\label{sec:dataset}

\begin{figure*}
\centering
\begin{subfigure}{.7\textwidth}
  \centering
  \includegraphics[width=.9\linewidth]{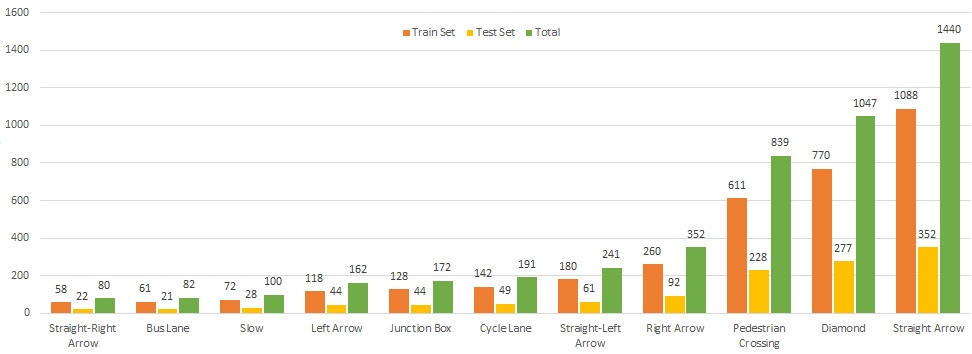}
  \vspace{-2mm}
  \caption{}
  \label{fig:datastat_class}
\end{subfigure}%
\begin{subfigure}{.3\textwidth}
  \centering
      \vspace{3mm}
  \includegraphics[width=\linewidth]{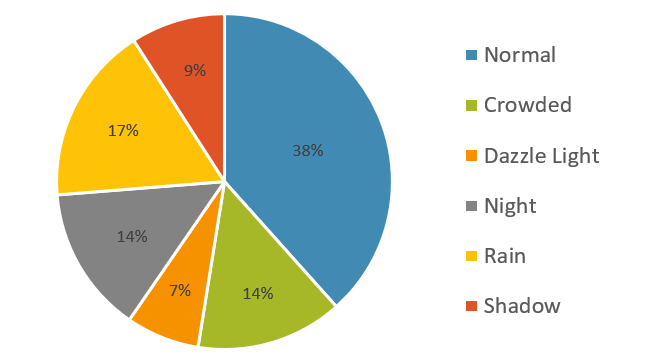}
    \vspace{-1mm}
  \caption{}
  \label{fig:datastat_scene}
\end{subfigure}
\caption{(a) Frequency of each class in the dataset. (b) Proportion of each scenario in the test set.}
\label{fig:datastat}
\end{figure*}

In this section, we present the CeyMo road marking benchmark dataset including the data collection and annotation processes, dataset statistics and the evaluation metrics.

\subsection{Data Collection}

Modern traffic sign and road marking datasets are created using footage from vehicles with specifically mounted cameras \cite{BDD, Cordts2016Cityscapes} or frames taken from street view services such as Google or Tencent \cite{Zhe_2016_CVPR}. We collect video footage from two cameras mounted inside of four vehicles to capture a wide range of scenarios including urban, sub-urban and rural areas under different weather and lighting conditions. The frames which contain road markings are then extracted from the recorded video footage.

\subsection{Data Annotation}

Road markings belonging to 11 classes are manually annotated as polygons using the labelme \cite{labelme2016} annotation tool. Each image has a JSON file which contains the coordinates of the polygons enclosing the road markings present in that image. In addition to the polygon annotations in the JSON format, we also provide bounding box annotations in the XML format as well as pixel-level segmentation masks in the PNG format to facilitate different road marking detection approaches. The three annotation formats provided with the dataset are visualized in Figure \ref{fi:annotations}. 

\begin{table}
\begin{center}
\begin{tabular}{|l|c|c|c|}
\hline
\textbf{Scenario} & \textbf{No. of Images} &\textbf{Percentage}  \\
\hline\hline
Normal          & 303  & 38.45\%    \\ \hline
Crowded         & 109    & 13.83\%    \\ \hline
Dazzle Light    & 56    & 7.11\%     \\ \hline
Night           & 110    & 13.96\%    \\ \hline
Rain            & 136    & 17.26\%    \\ \hline
Shadow          & 74     & 9.39\%    \\ \hline

\end{tabular}
\end{center}
\vspace{-1em}
\caption{Number of images for each scenario in the test set.}
\vspace{-1em}
\label{tab:scenarios}
\end{table}

\subsection{Dataset Statistics}


Our new benchmark consists of 2887 total images having a resolution of $1920 \times 1080$. The dataset is divided into the train set and the test set, which comprises 2099 images and 788 images, respectively. The benchmark covers 11 different road marking classes and the number of instances present in each class is shown in Table \ref{tab:instances}. There is an inherent class imbalance in the dataset which is highlighted in Figure \ref{fig:datastat_class}.

The 788 images in the test set are divided into 6 categories including normal and 5 challenging scenarios: crowded, dazzle light, night, rain and shadow. The number of images and the proportion of each category are shown in Table \ref{tab:scenarios} and Figure \ref{fig:datastat_scene}. It can be observed that the 5 challenging scenarios account for the majority (61.55\%) of the test set.


\begin{table}
\begin{center}
\begin{tabular}{|l|c|c|c|c|}
\hline
\textbf{Road Marking Class} & \textbf{Train Set} & \textbf{Test Set} & \textbf{Total} \\
\hline\hline
Straight   Arrow                         & 1088          & 352          & 1440             \\ \hline
Left   Arrow                             & 118           & 44           & 162              \\ \hline
Right   Arrow                            & 260           & 92           & 352              \\ \hline
Straight-Left   Arrow                    & 180           & 61           & 241              \\ \hline
Straight-Right   Arrow                   & 58            & 22           & 80               \\ \hline
Diamond                                  & 770           & 277          & 1047              \\ \hline
Pedestrian   Crossing                    & 611           & 228          & 839               \\ \hline
Junction   Box                           & 128           & 44           & 172               \\ \hline
Slow                                     & 72            & 28           & 100               \\ \hline
Bus Lane                                 & 61            & 21           & 82                \\ \hline
Cycle   Lane                             & 142           & 49           & 191               \\ \hline
\textbf{Total}      & \textbf{3488}  & \textbf{1218} & \textbf{4706}           \\ \hline
\end{tabular}
\end{center}
\vspace{-1em}
\caption{Number of instances for each class in the dataset.}
\vspace{-1em}
\label{tab:instances}
\end{table}





\begin{figure*}
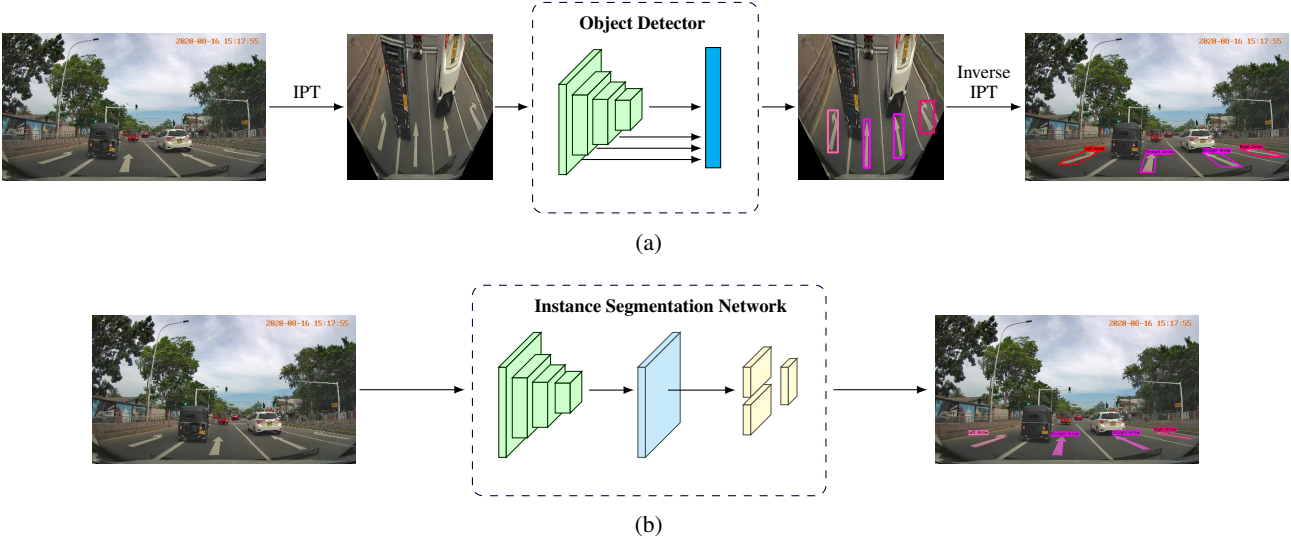

     \centering
     \begin{subfigure}[b]{1\linewidth}
         \centering
         \input{arch}
         \caption{}
         \label{roadarch:sf1}
         \vspace{2ex}
     \end{subfigure}%
     
     \begin{subfigure}[b]{1\linewidth}
         \centering
         \input{arch_mask}
         \caption{}
         \label{roadarch:sf2}
     \end{subfigure}%
    \caption{ (a) Proposed object detection based network architecture. The inverse perspective transform is used to obtain the bird's eye view of the road which will be fed to the object detector for detecting road markings as bounding boxes. The detections are mapped to 4-sided polygons in the original image using the inverse of the IPT matrix. SSD-MobileNet-v1 \cite{SSD,mobilenet} and SSD-Inception-v2 \cite{SSD,inception} are evaluated as the object detectors. (b) Proposed instance segmentation based network architecture. Two Mask R-CNN \cite{mask_rcnn} based networks with Inception-v2 \cite{inception} and ResNet-50 \cite{resnet} backbones are evaluated for detecting road markings on the input images as segmentation masks.}
    
    \label{fi:roadarch}
    \vspace{-0.7em}
\end{figure*}

\subsection{Evaluation Metrics}

 We use $F_{1}$-score and Macro $F_{1}$-score as the evaluation metrics of our road marking dataset. Intersection over union values between the predictions and the ground truth are calculated and if IoU is greater than 0.3, the corresponding prediction is considered as a true positive. The total number of true positives ($TP$), false positives ($FP$) and false negatives ($FN$) are used to calculate the precision, recall and $F_{1}$-measure as follows:
 
\begin{equation}
precision = \frac{TP}{TP+FP}
\end{equation}

\begin{equation}
recall = \frac{TP}{TP+FN}
\end{equation}

\begin{equation}
    F_{1}\mbox{-}score= \frac{2\times precision \times recall}{precision+recall}
\end{equation}

\vspace{0.5em}

Macro $F_{1}$-score is calculated as the mean of the individual $F_{1}$-scores of the 11 classes present in our dataset as follows: 

\vspace{-1em}

\begin{equation}
    Macro\mbox{-}F_{1}\mbox{-}score =\dfrac{1}{C} \sum_{i=1}^{C} F_{1}\mbox{-}score_{i} 
\end{equation}

Macro $F_{1}$-score gives same importance for all classes regardless of the frequencies they appear in the dataset. Therefore, it will be low for models which only perform well on common classes.


The evaluation script which calculates class-wise, scenario-wise and overall precision, recall and $F_{1}$-score values, and the Macro $F_{1}$-score, will be made publicly available facilitating direct comparison of different road marking detection algorithms.


\section{Methodology}
\label{sec:method}



In this section, we explain our two detection pipelines under object detection and instance segmentation based approaches, that are used to detect road markings on the CeyMo road marking dataset.

\begin{table*}[t!]
\small
\begin{subtable}{\textwidth}
\begin{center}
\begin{tabular}{|p{3cm}|c|c|c|c|}
\hline

\textbf{Category} & \textbf{SSD-MobileNet-v1} & \textbf{SSD-Inception-v2} & \textbf{Mask-RCNN-Inception-v2} & \textbf{Mask-RCNN-ResNet50} \\

& \cite{SSD,mobilenet} & \cite{SSD,inception} & \cite{mask_rcnn,inception} & \cite{mask_rcnn,resnet} \\

\hline\hline
Normal                             & 86.57                   & 87.10                   & 93.20                         & \textbf{94.14}                       \\ \hline
Crowded                            & 79.45                   & 82.51                  & 82.04                        & \textbf{85.78}                       \\ \hline
Dazzle light                       & 84.97                   & 85.90                   & 86.06                        & \textbf{89.29}                       \\ \hline
Night                              & 83.08                   & 84.85                  & \textbf{92.59}                        & 91.51                       \\ \hline
Rain                               & 73.68                   & 81.87                  & 87.50                         & \textbf{89.08}                       \\ \hline
Shadow                             & 85.25                   & 86.53                  & 85.60                         & \textbf{87.30}                        \\ \hline
\textbf{Total}              & 82.90                   & 85.16                  & 89.04                        & \textbf{90.62}              \\ \hline
\textbf{Speed (FPS)}               & \textbf{83}             & 61                    & 42                          & 13                         \\ \hline
\end{tabular}

\caption{}
\label{tab:results_scenario}
\end{center}
\end{subtable}

\vspace{1em}

\begin{subtable}{\textwidth}
\begin{center}
\begin{tabular}{|p{3cm}|c|c|c|c|}
\hline
\textbf{Class} & \textbf{SSD-MobileNet-v1} & \textbf{SSD-Inception-v2} & \textbf{Mask-RCNN-Inception-v2} & \textbf{Mask-RCNN-ResNet50} \\

& \cite{SSD,mobilenet} & \cite{SSD,inception} & \cite{mask_rcnn,inception} & \cite{mask_rcnn,resnet} \\

\hline\hline
Straight Arrow       & 73.51         & 77.39         & 86.00                  & 88.33              \\ \hline
Left Arrow           & 66.67         & 73.97         & 59.70                & 74.36              \\ \hline
Right Arrow          & 75.64         & 81.93         & 84.75               & 90.40               \\ \hline
Straight-Left Arrow  & 65.22         & 65.93         & 84.55               & 89.47              \\ \hline
Straight-Right Arrow & 62.50          & 58.06         & 74.29               & 66.67              \\ \hline
Diamond              & 87.82         & 88.58         & 92.05               & 91.05              \\ \hline
Pedestrian Crossing  & 94.95         & 95.44         & 96.72               & 96.86              \\ \hline
Junction Box         & 82.50          & 90.70          & 92.13               & 96.63              \\ \hline
Slow                 & 88.46         & 90.20          & 92.59               & 94.34              \\ \hline
Bus Lane             & 98.00            & 100.00           & 93.33               & 91.26              \\ \hline
Cycle Lane           & 95.00            & 89.47         & 87.18               & 92.31              \\ \hline
\textbf{Macro F1-Score}& 80.93         & 82.88         & 85.75               & \textbf{88.33}              \\ \hline
\end{tabular}
\caption{}
\label{tab:results_class}
\end{center}
\end{subtable}
\caption{Road marking detection results. (a) For each of the detection models scenario-wise F1-scores, overall F1-score, and inference speed in frames per second (FPS) are listed. The inference speed is measured by taking the average FPS value for the 788 test images. (b) For each of the detection models class-wise F1-scores and the Macro F1-score are listed.}
\vspace{-1em}
\end{table*}

\subsection{Object Detection Approach}

Our object detection based model architecture for the road marking detection task is shown in Figure \ref{roadarch:sf1}. Each image is first transformed using the inverse perspective transform (IPT) to obtain a bird's eye view of the road area. IPT reduces perspective deformation of the captured images and it also removes a larger area of the background and the road markings become more prominent in the resultant image. The inverse perspective transform is a homography transformation given by the following equations, where $M$ is the relevant homography matrix.

\begin{equation}
\label{eq:IPT}
Destination[\hat{x},\hat{y},:] = Source[x,y,:]
\end{equation}
where,
\begin{equation}
\hat{x}=\frac{M_{11}x+M_{12}y+M_{13}}{M_{31}x+M_{32}y+M_{33}}
\end{equation}
\begin{equation}
\hat{y}=\frac{M_{21}x+M_{22}y+M_{23}}{M_{31}x+M_{32}y+M_{33}}
\end{equation}

\vspace{0.5em}

 An end-to-end object detector model is then used to detect road markings on the inverse perspective transformed images. We evaluate the performance of two object detector models for the road marking detection task,
 SSD \cite{SSD} with MobileNet-v1 backbone \cite{ mobilenet} and SSD \cite{SSD} with Inception-v2 \cite{inception} as the backbone. The input resolution of both the models is set to $ 500 \times 500 $. These object detector models output the road marking detections as bounding boxes on the inverse perspective transformed image. These bounding box detections are transformed to the original image domain as 4-sided polygons using the inverse of the IPT homography matrix ($M$) as the final step.
 
 \subsection{Instance Segmentation Approach}

Figure \ref{roadarch:sf2} depicts the model architecture used for road marking detection under the instance segmentation based approach. The goal of instance segmentation is to predict object instances and their per-pixel segmentation masks. For this task, we employ the widely used Mask R-CNN \cite{mask_rcnn} network architecture with two backbones, Inception-v2 \cite{inception} and ResNet-50 \cite{resnet}. Mask R-CNN \cite{mask_rcnn} extends the Faster R-CNN \cite{FRCNN} architecture by predicting a segmentation mask, in addition to the bounding box, for each region of interest (RoI) identified. 

Since instance segmentation networks usually have low inference speeds, we feed the input images directly to the model, after resizing them into a lower resolution of $500 \times 500$, without any additional pre-processing steps. The network outputs both bounding boxes and segmentation masks for the road marking detections. Yet, our interest only lies in the segmentation masks, from which the convex hulls could be obtained for evaluation purposes.





\section{Experiments}
\label{sec:Experiments}

\begin{figure*}
     \centering
     \begin{subfigure}[b]{0.16\linewidth}
         \centering
         \includegraphics[width=.98\linewidth]{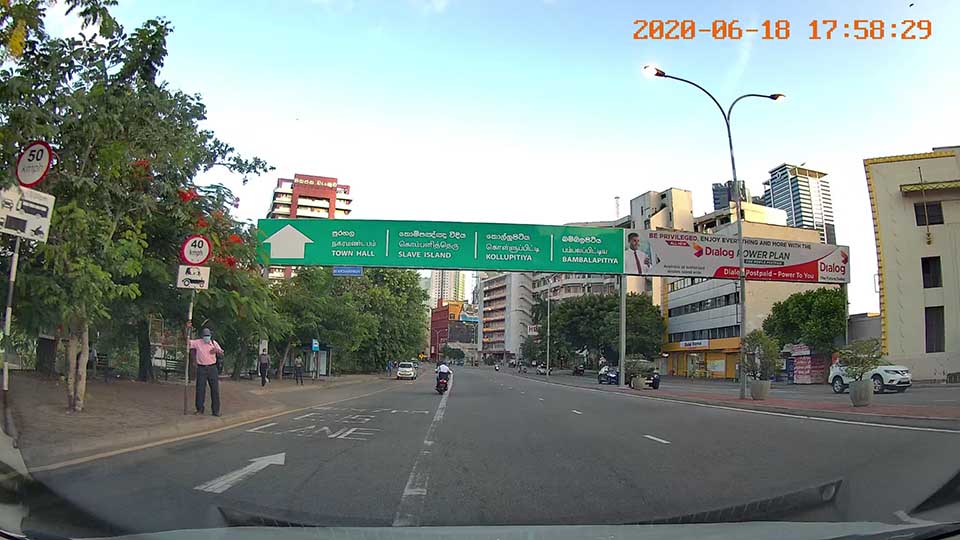}
     \end{subfigure}%
     \begin{subfigure}[b]{0.16\linewidth}
         \centering
         \includegraphics[width=.98\linewidth]{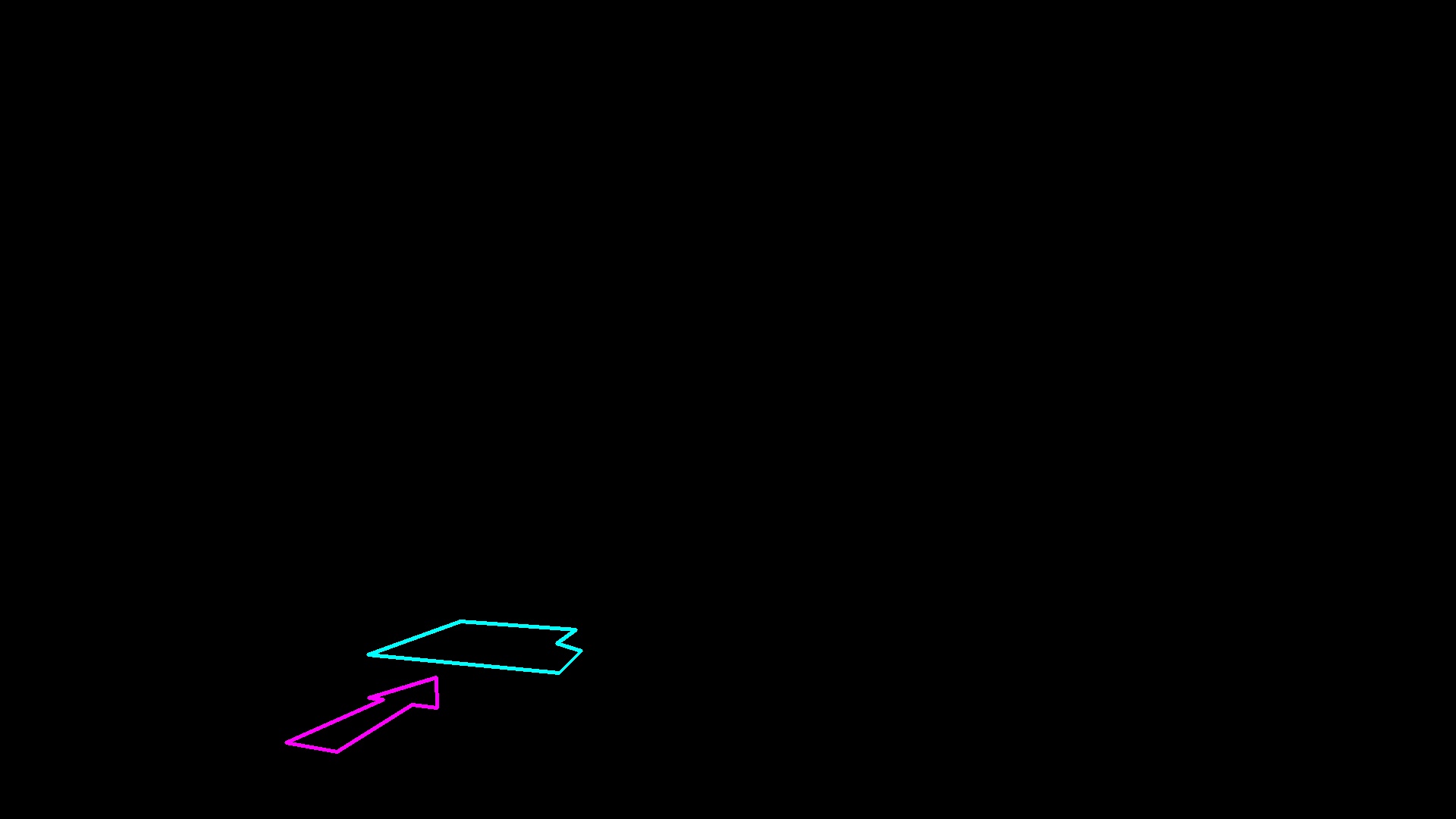}
     \end{subfigure}%
     \begin{subfigure}[b]{0.16\linewidth}
         \centering
         \includegraphics[width=.98\linewidth]{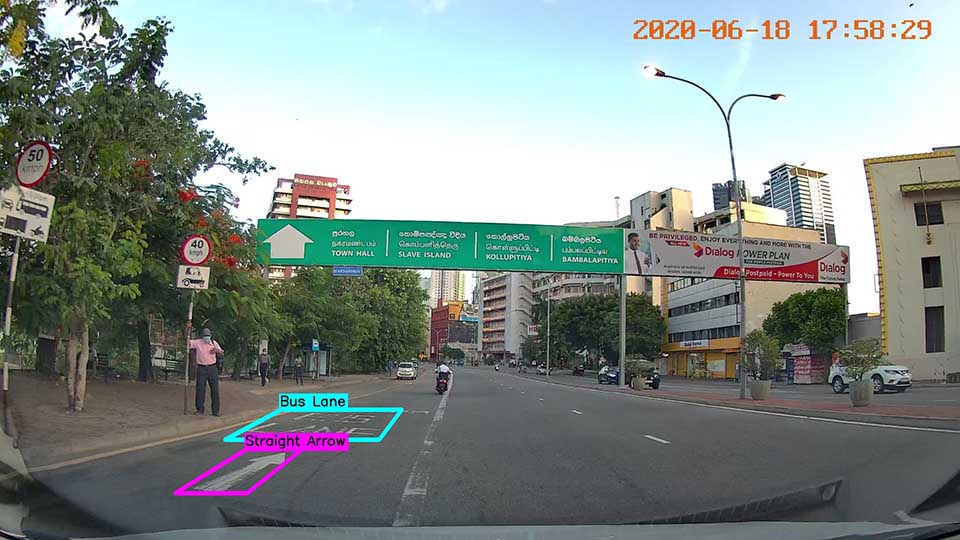}
     \end{subfigure}%
     \begin{subfigure}[b]{0.16\linewidth}
         \centering
         \includegraphics[width=.98\linewidth]{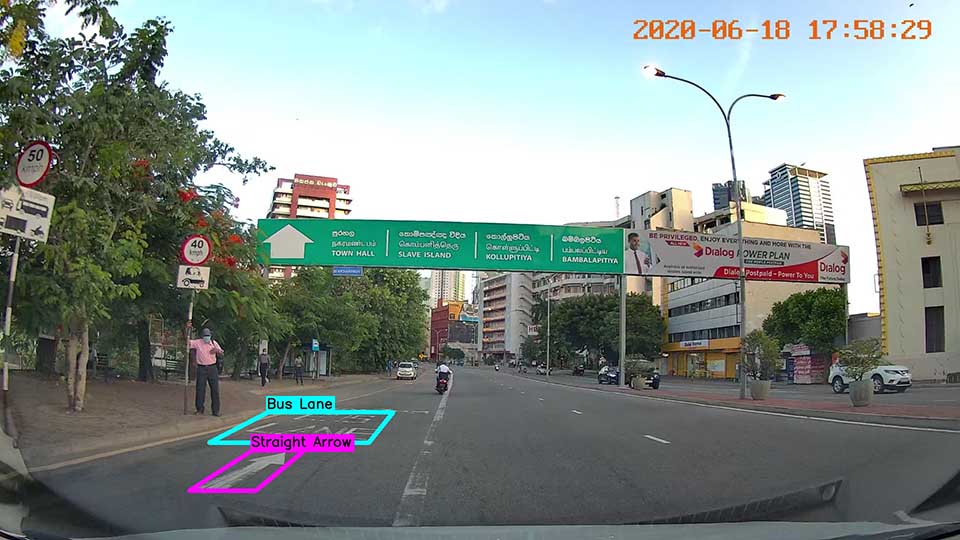}
     \end{subfigure}%
     \begin{subfigure}[b]{0.16\linewidth}
         \centering
         \includegraphics[width=.98\linewidth]{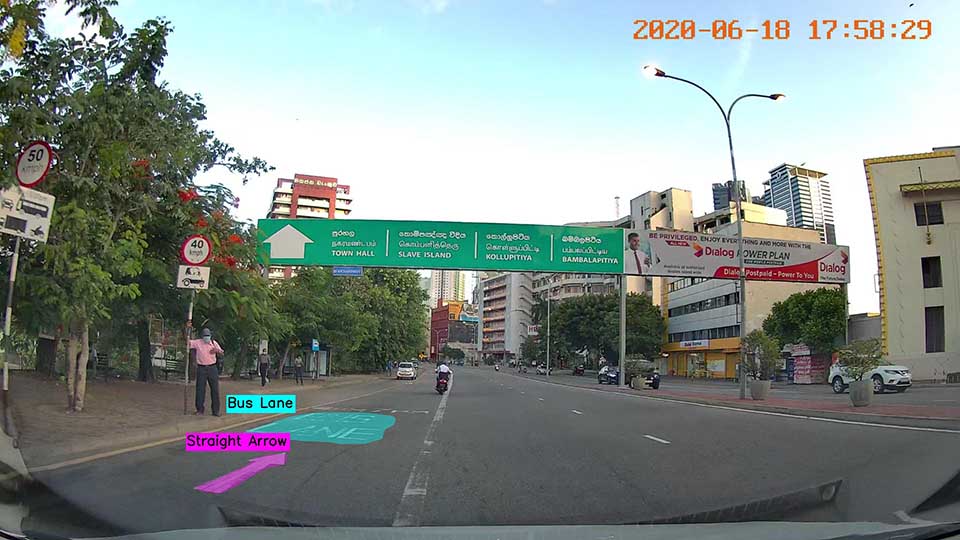}
     \end{subfigure}%
     \begin{subfigure}[b]{0.16\linewidth}
         \centering
         \includegraphics[width=.98\linewidth]{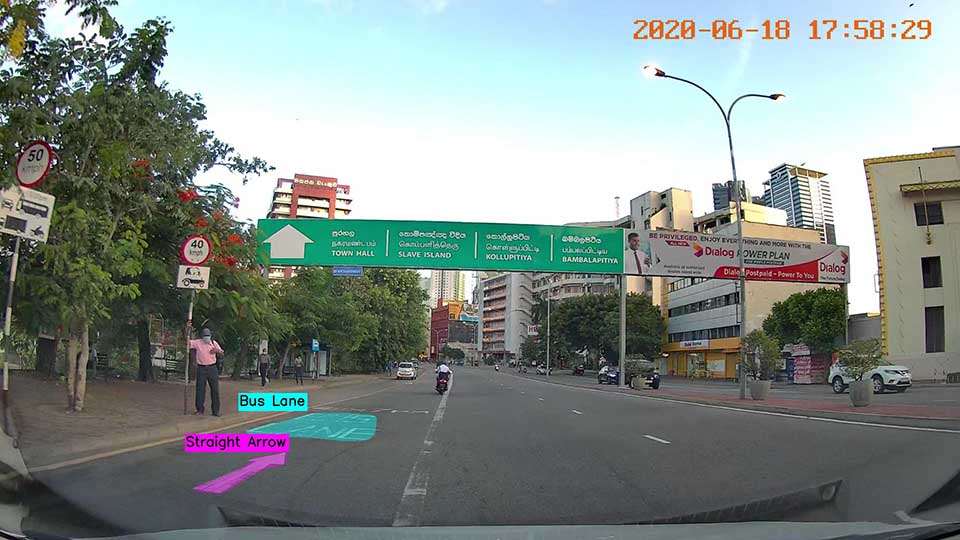}
     \end{subfigure}%
     
     \begin{subfigure}[b]{0.16\linewidth}
         \centering
         \includegraphics[width=.98\linewidth]{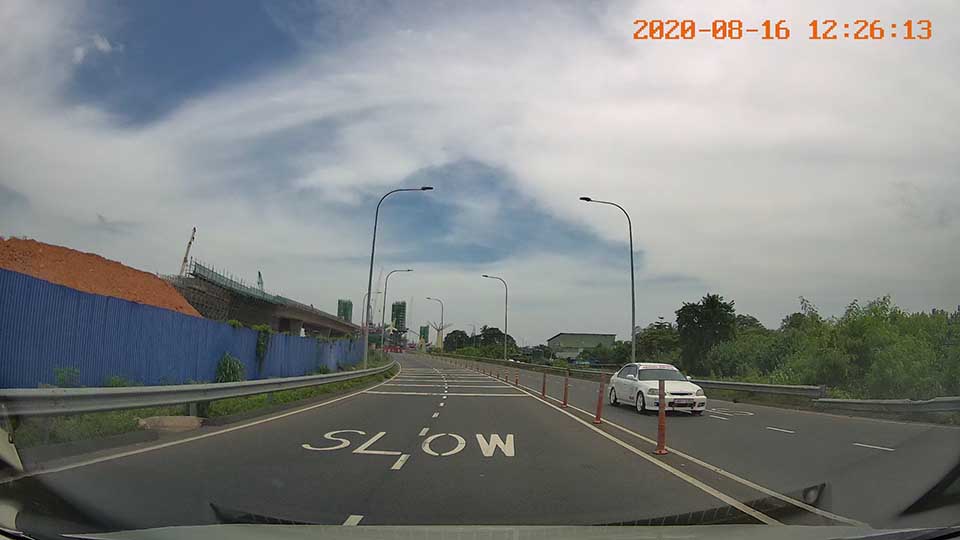}
     \end{subfigure}%
     \begin{subfigure}[b]{0.16\linewidth}
         \centering
         \includegraphics[width=.98\linewidth]{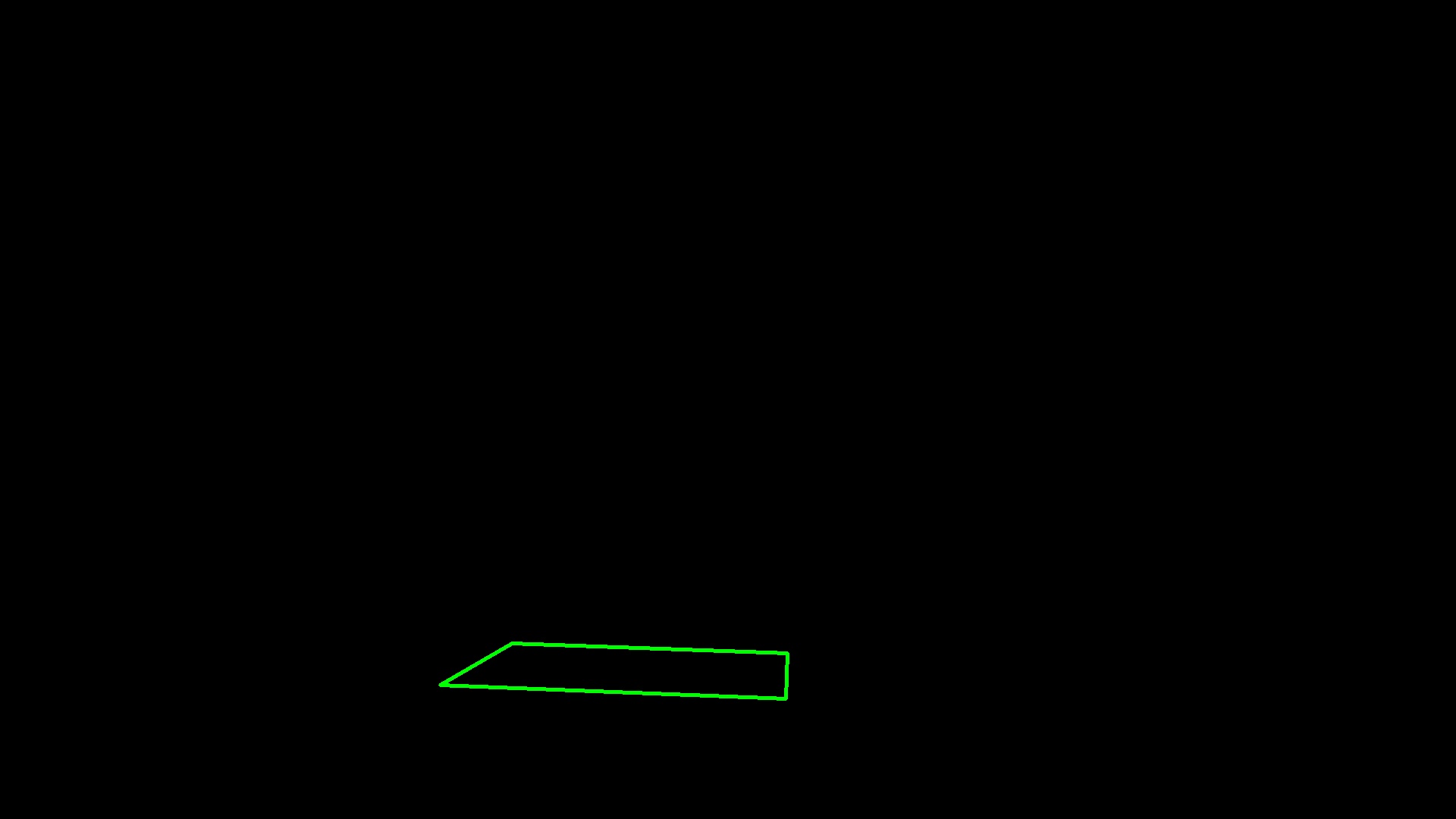}
     \end{subfigure}%
     \begin{subfigure}[b]{0.16\linewidth}
         \centering
         \includegraphics[width=.98\linewidth]{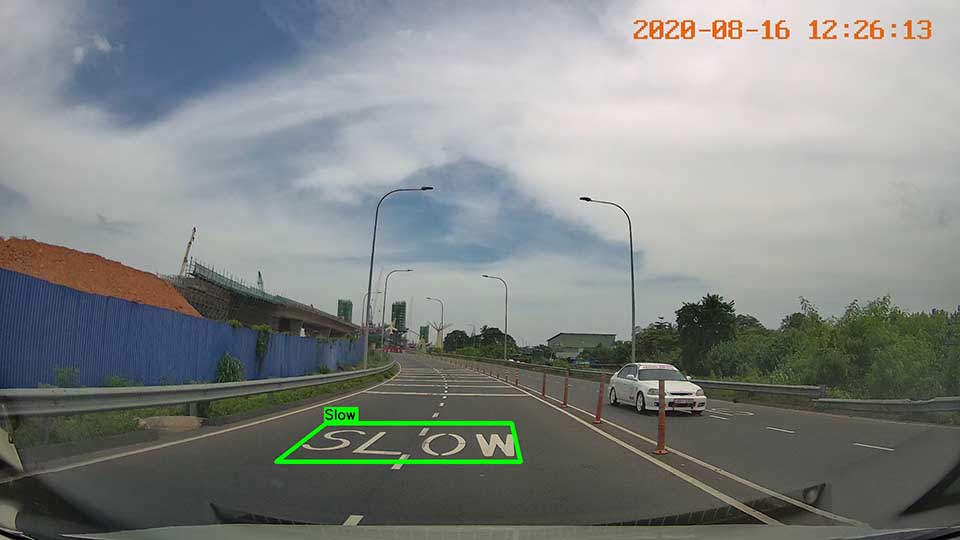}
     \end{subfigure}%
     \begin{subfigure}[b]{0.16\linewidth}
         \centering
         \includegraphics[width=.98\linewidth]{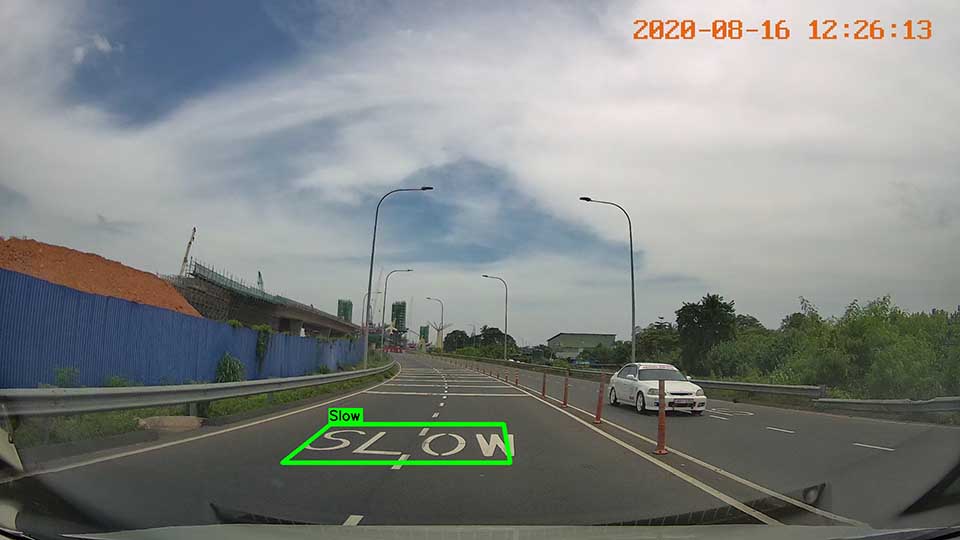}
     \end{subfigure}%
     \begin{subfigure}[b]{0.16\linewidth}
         \centering
         \includegraphics[width=.98\linewidth]{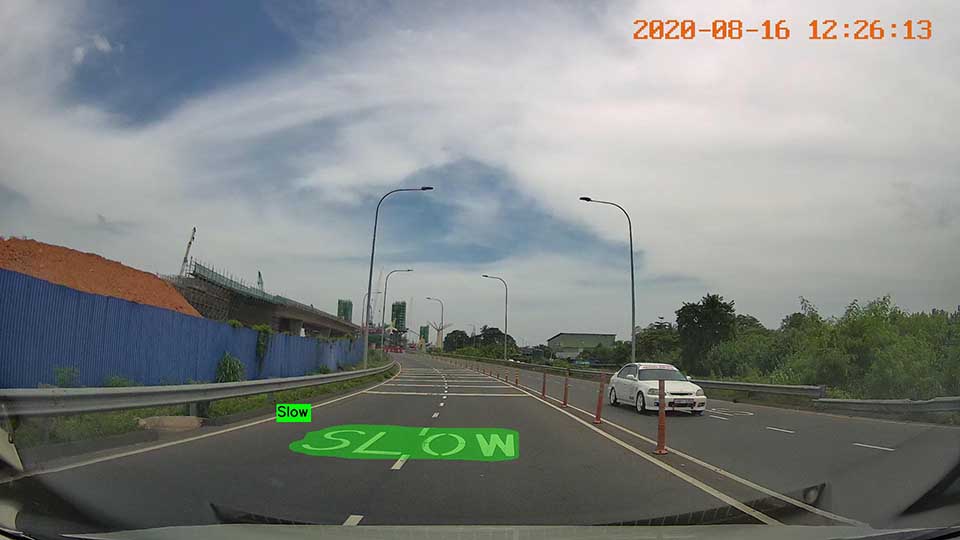}
     \end{subfigure}%
     \begin{subfigure}[b]{0.16\linewidth}
         \centering
         \includegraphics[width=.98\linewidth]{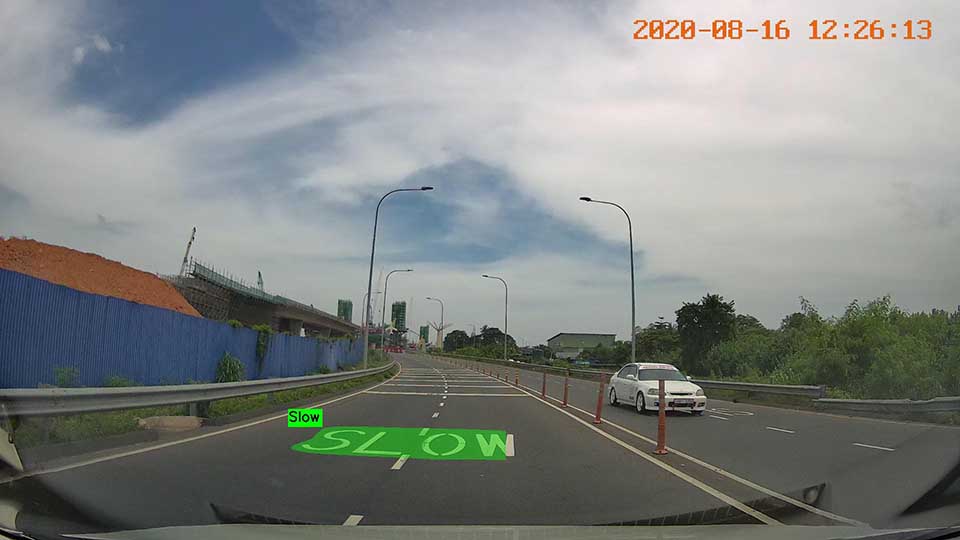}
     \end{subfigure}%
     
     \begin{subfigure}[b]{0.16\linewidth}
         \centering
         \includegraphics[width=.98\linewidth]{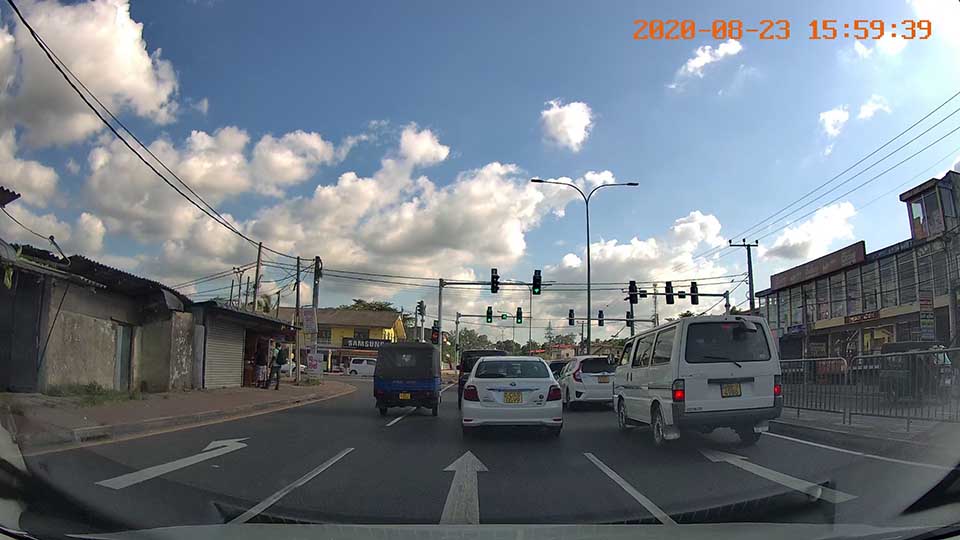}
     \end{subfigure}%
     \begin{subfigure}[b]{0.16\linewidth}
         \centering
         \includegraphics[width=.98\linewidth]{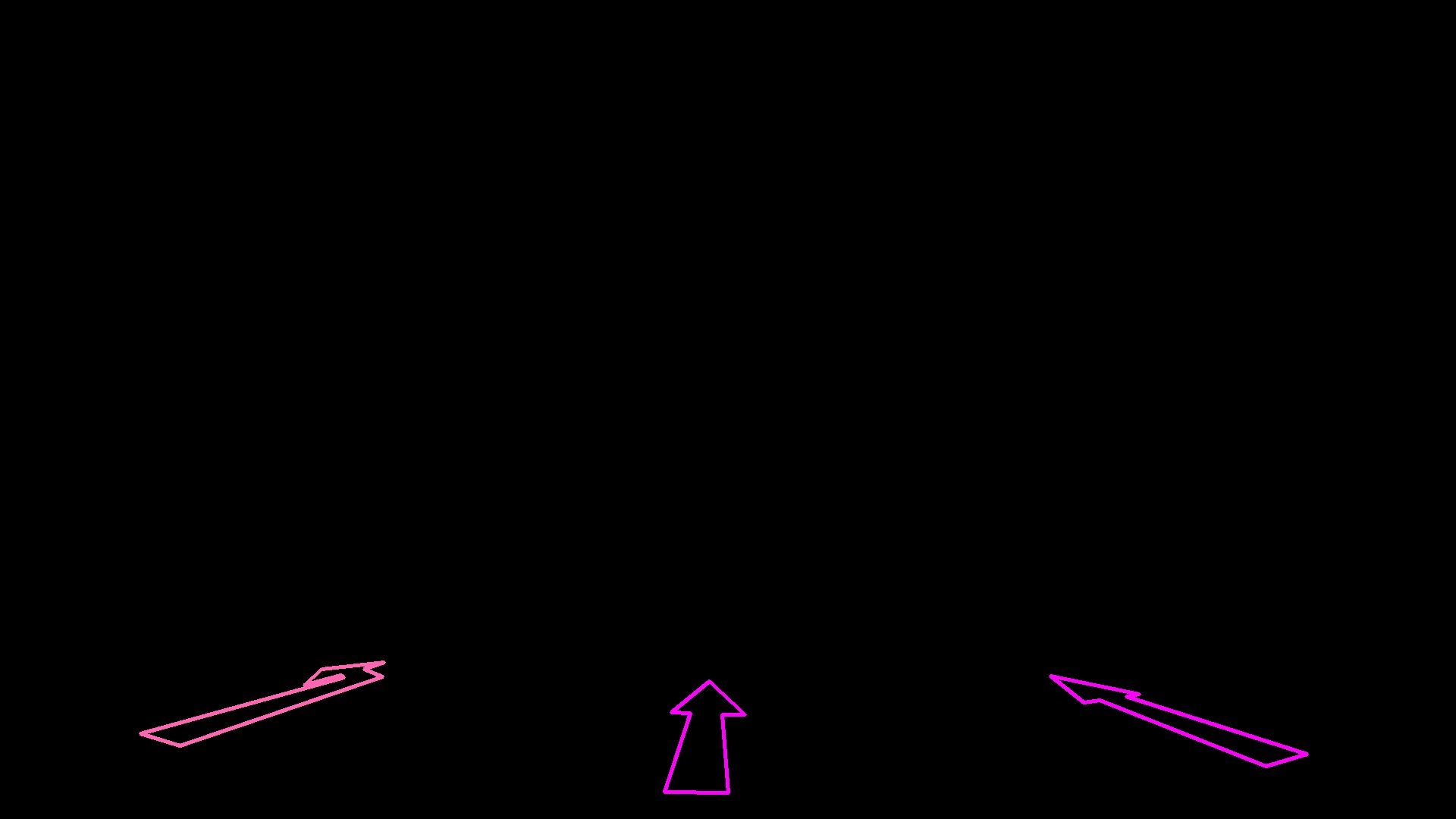}
     \end{subfigure}%
     \begin{subfigure}[b]{0.16\linewidth}
         \centering
         \includegraphics[width=.98\linewidth]{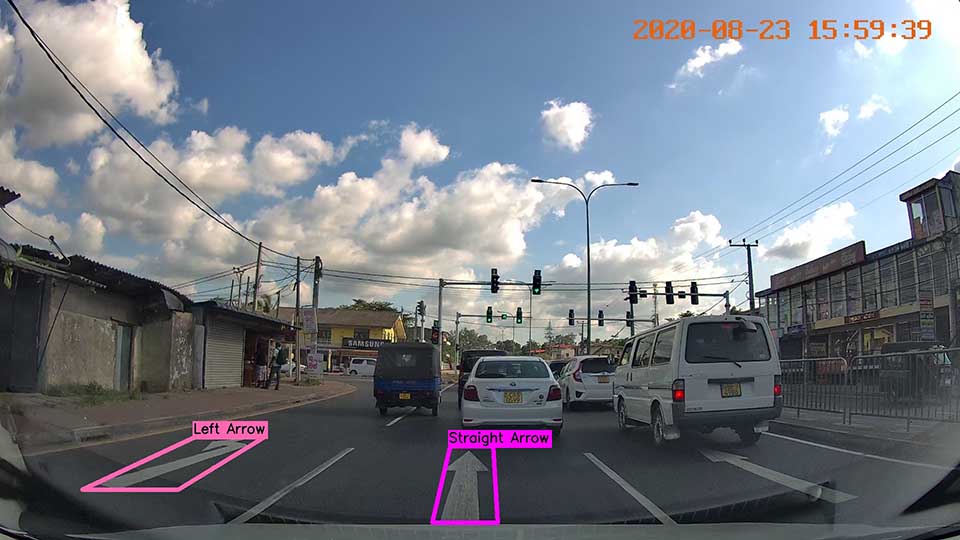}
     \end{subfigure}%
     \begin{subfigure}[b]{0.16\linewidth}
         \centering
         \includegraphics[width=.98\linewidth]{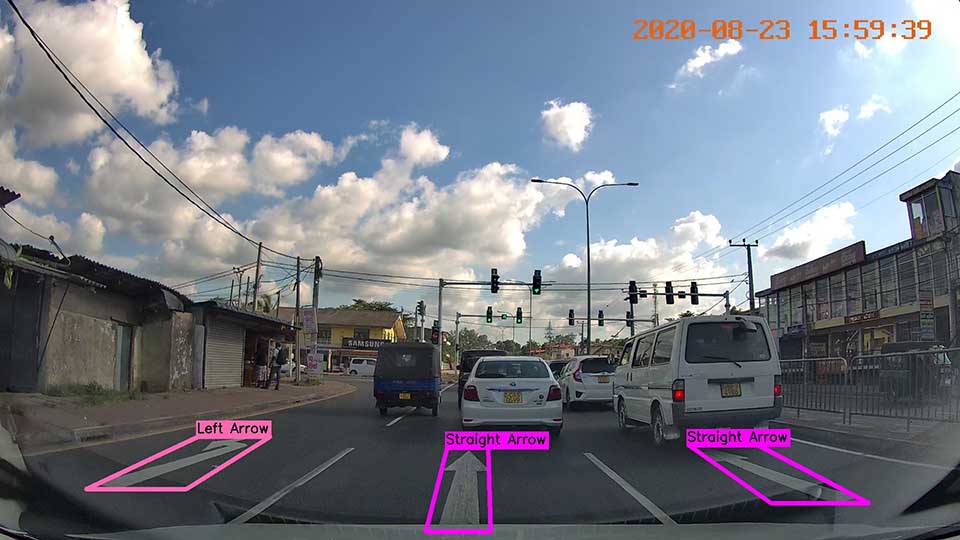}
     \end{subfigure}%
     \begin{subfigure}[b]{0.16\linewidth}
         \centering
         \includegraphics[width=.98\linewidth]{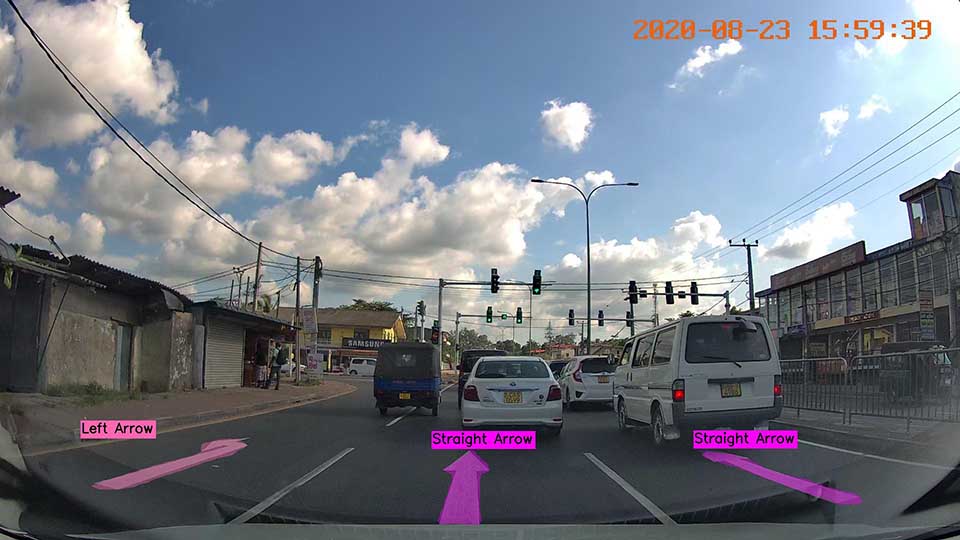}
     \end{subfigure}%
     \begin{subfigure}[b]{0.16\linewidth}
         \centering
         \includegraphics[width=.98\linewidth]{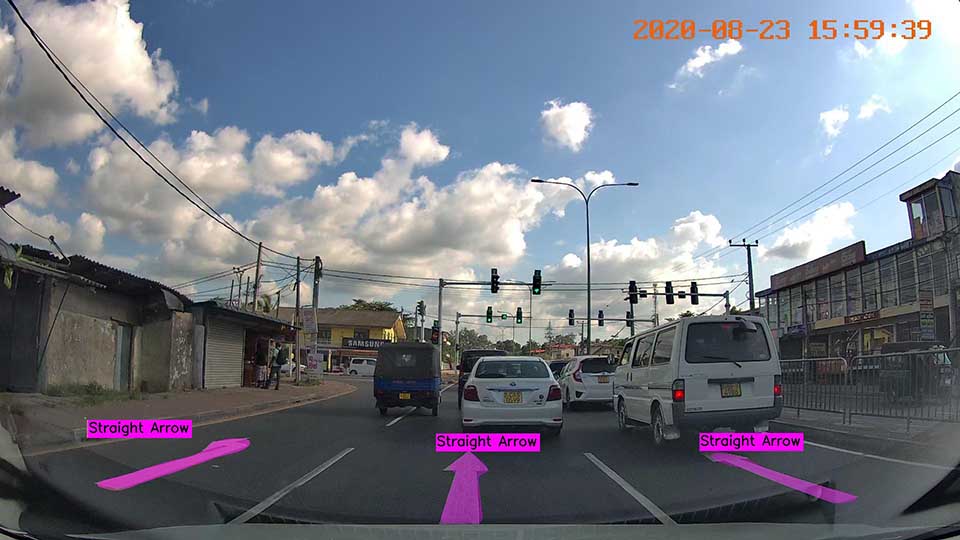}
     \end{subfigure}%
     
     \begin{subfigure}[b]{0.16\linewidth}
         \centering
         \includegraphics[width=.98\linewidth]{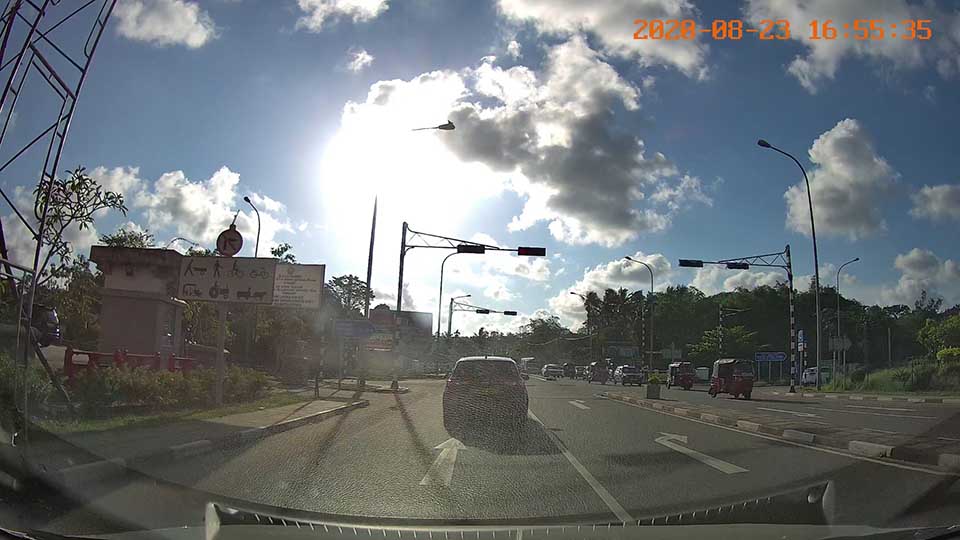}
     \end{subfigure}%
     \begin{subfigure}[b]{0.16\linewidth}
         \centering
         \includegraphics[width=.98\linewidth]{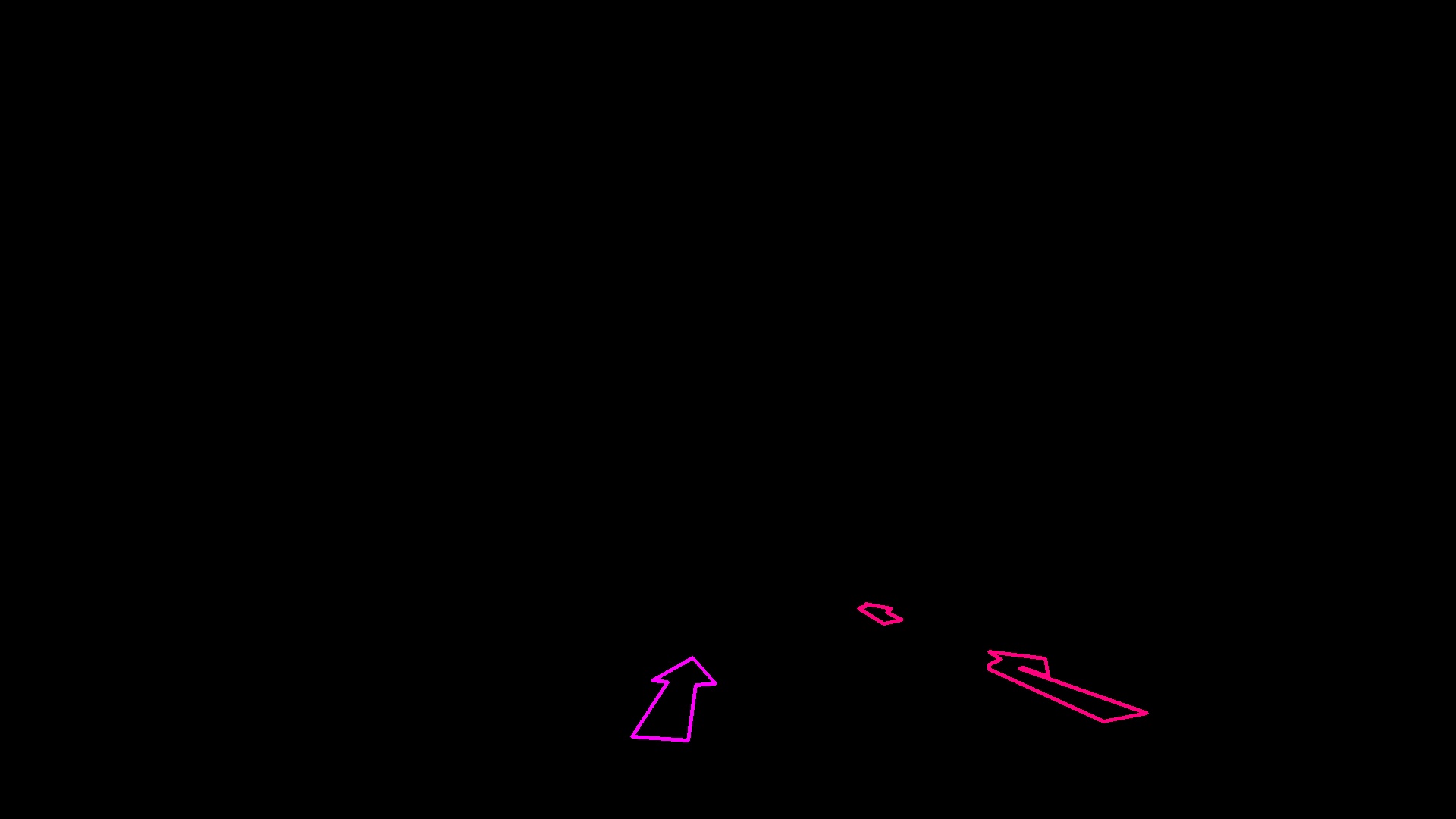}
     \end{subfigure}%
     \begin{subfigure}[b]{0.16\linewidth}
         \centering
         \includegraphics[width=.98\linewidth]{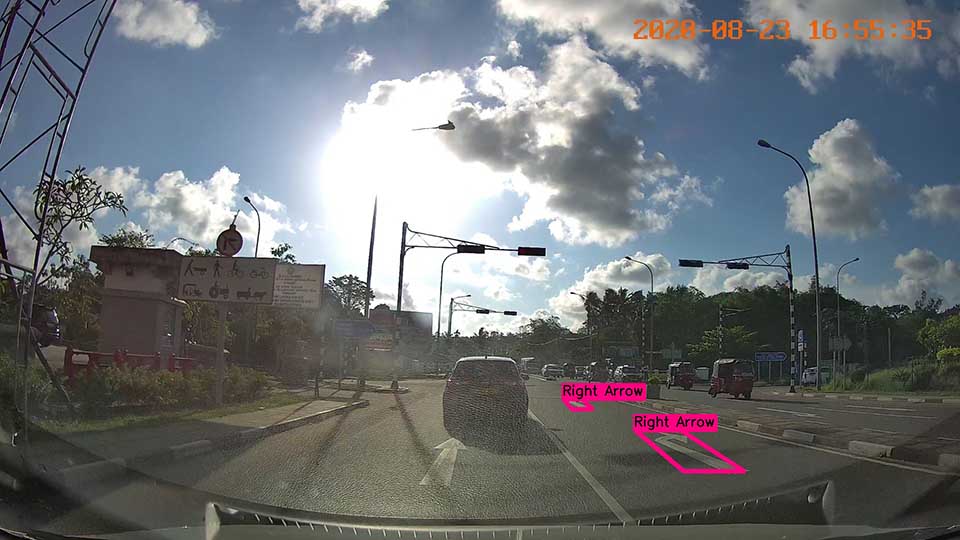}
     \end{subfigure}%
     \begin{subfigure}[b]{0.16\linewidth}
         \centering
         \includegraphics[width=.98\linewidth]{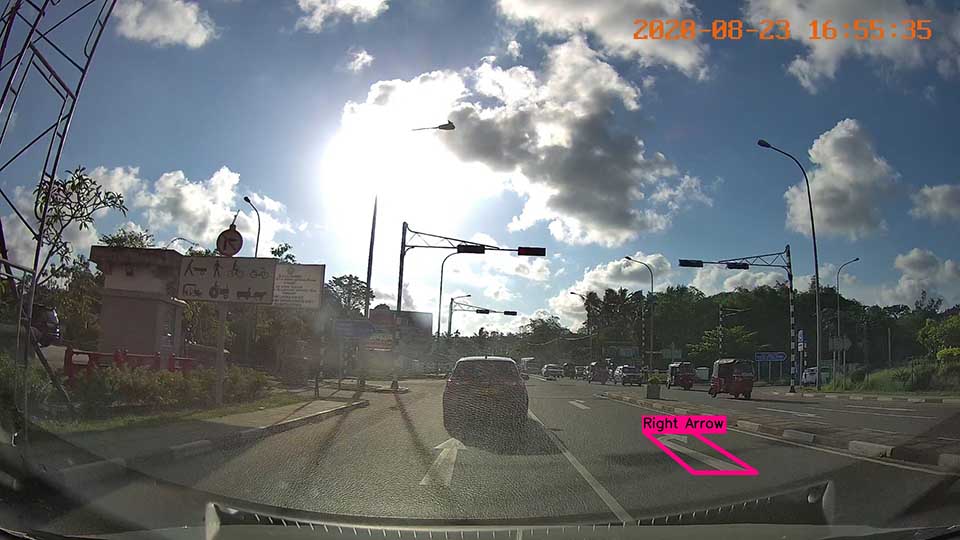}
     \end{subfigure}%
     \begin{subfigure}[b]{0.16\linewidth}
         \centering
         \includegraphics[width=.98\linewidth]{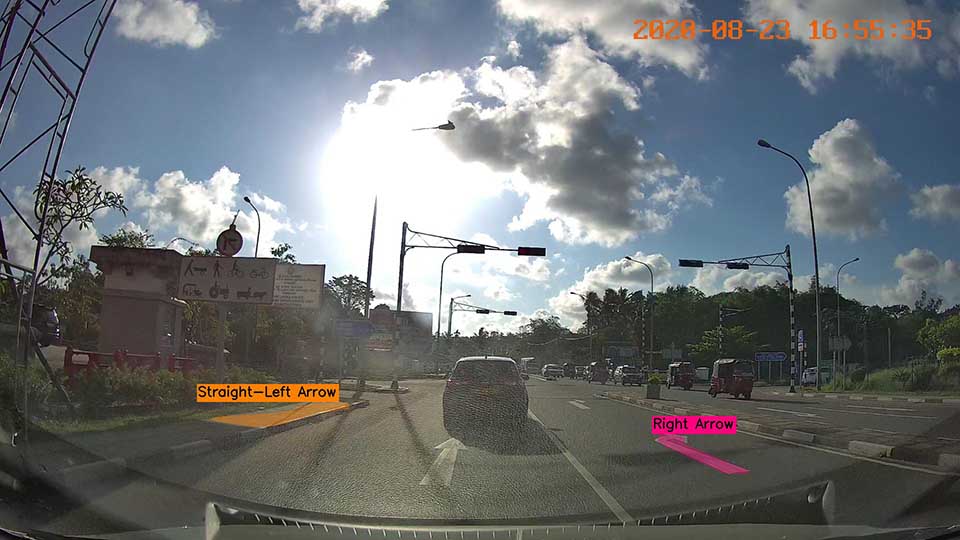}
     \end{subfigure}%
     \begin{subfigure}[b]{0.16\linewidth}
         \centering
         \includegraphics[width=.98\linewidth]{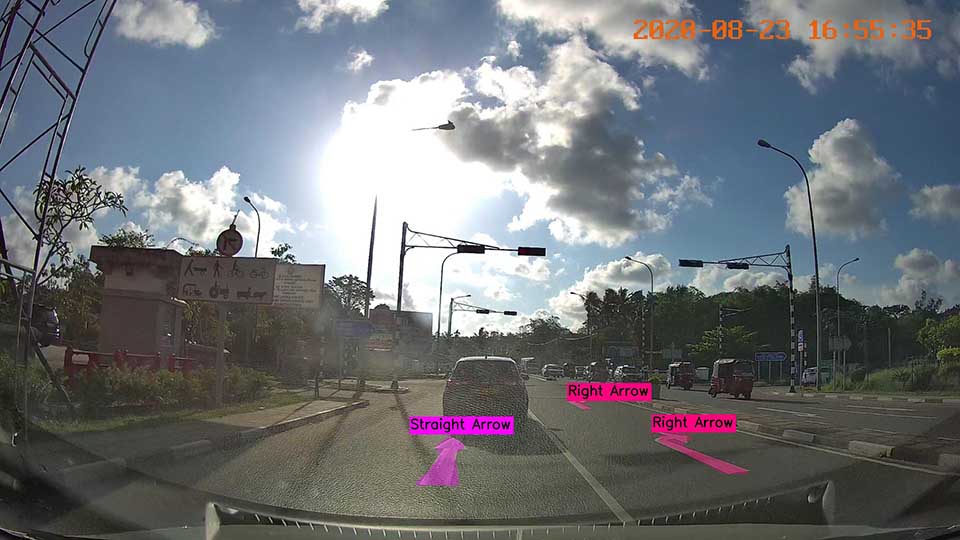}
     \end{subfigure}%
     
     \begin{subfigure}[b]{0.16\linewidth}
         \centering
         \includegraphics[width=.98\linewidth]{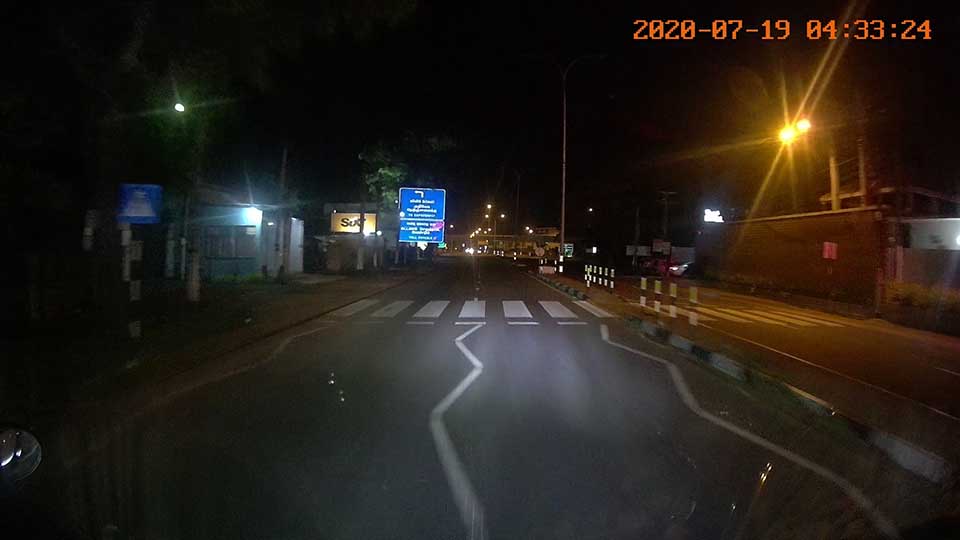}
     \end{subfigure}%
     \begin{subfigure}[b]{0.16\linewidth}
         \centering
         \includegraphics[width=.98\linewidth]{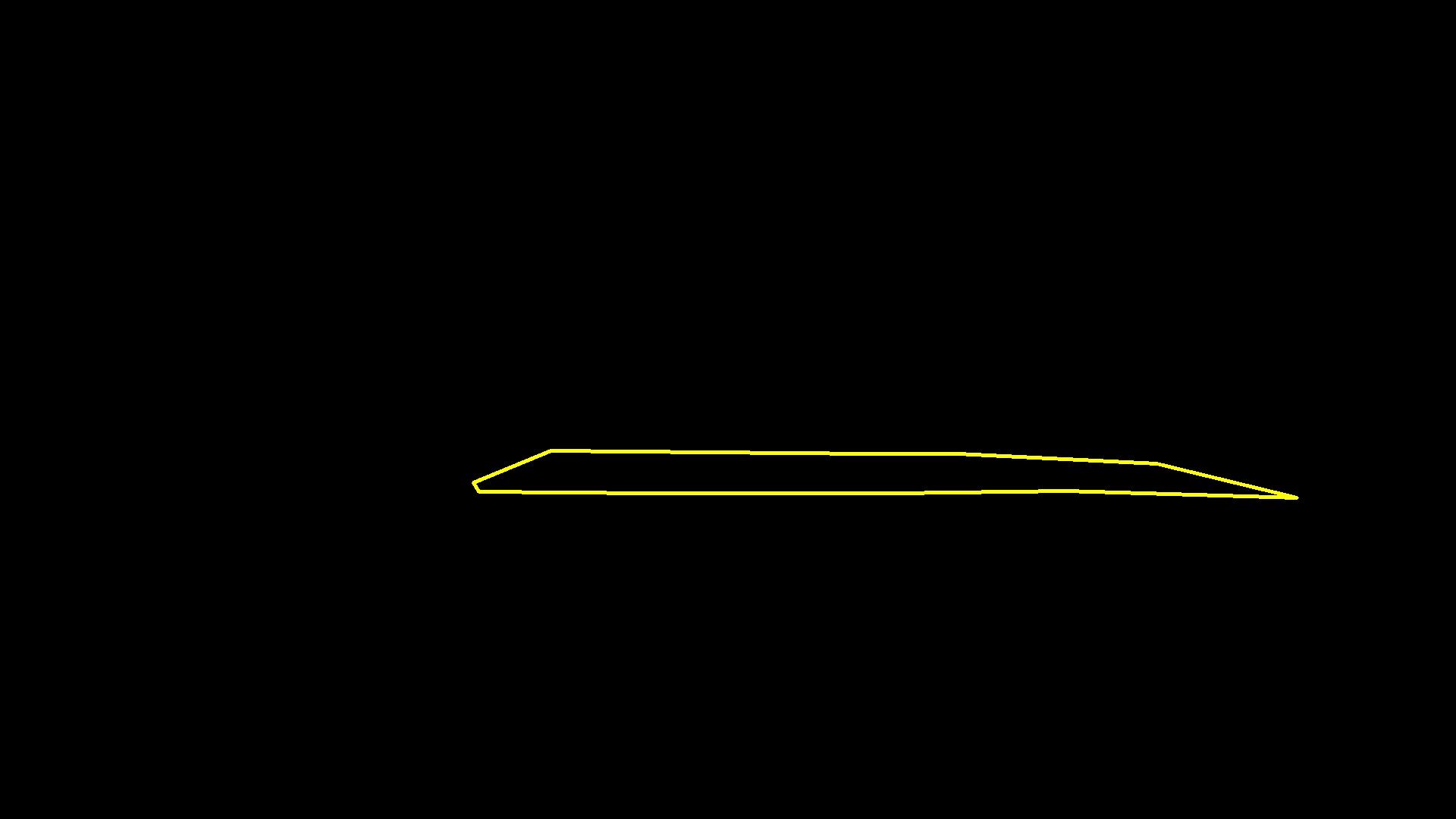}
     \end{subfigure}%
     \begin{subfigure}[b]{0.16\linewidth}
         \centering
         \includegraphics[width=.98\linewidth]{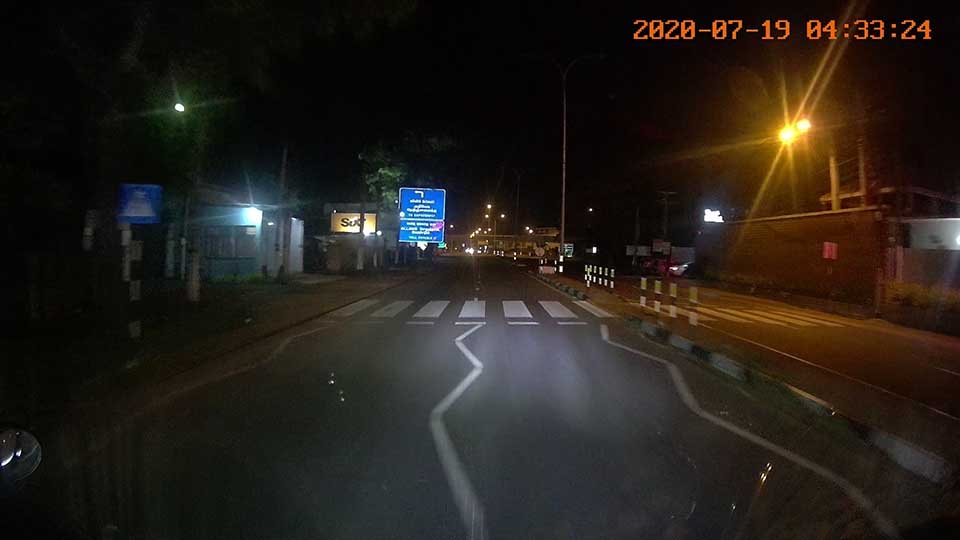}
     \end{subfigure}%
     \begin{subfigure}[b]{0.16\linewidth}
         \centering
         \includegraphics[width=.98\linewidth]{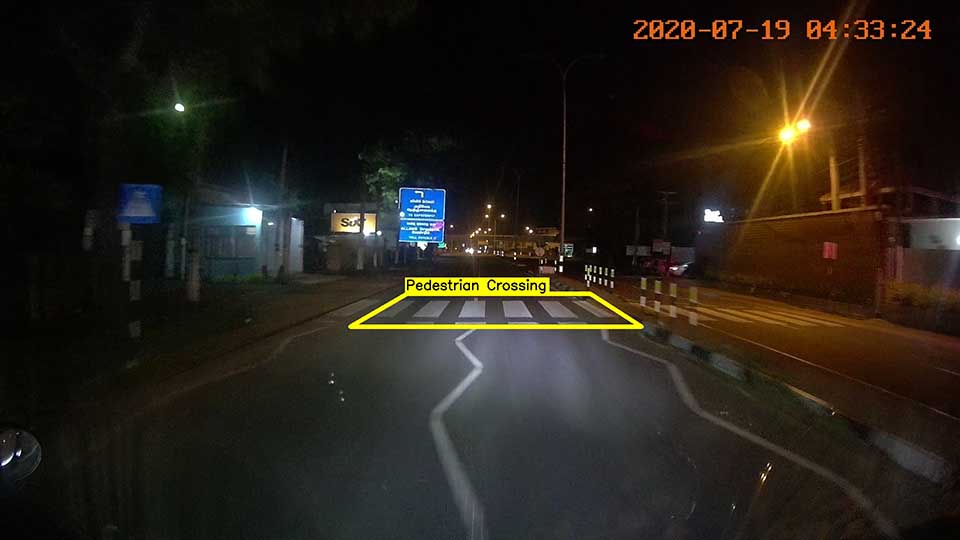}
     \end{subfigure}%
     \begin{subfigure}[b]{0.16\linewidth}
         \centering
         \includegraphics[width=.98\linewidth]{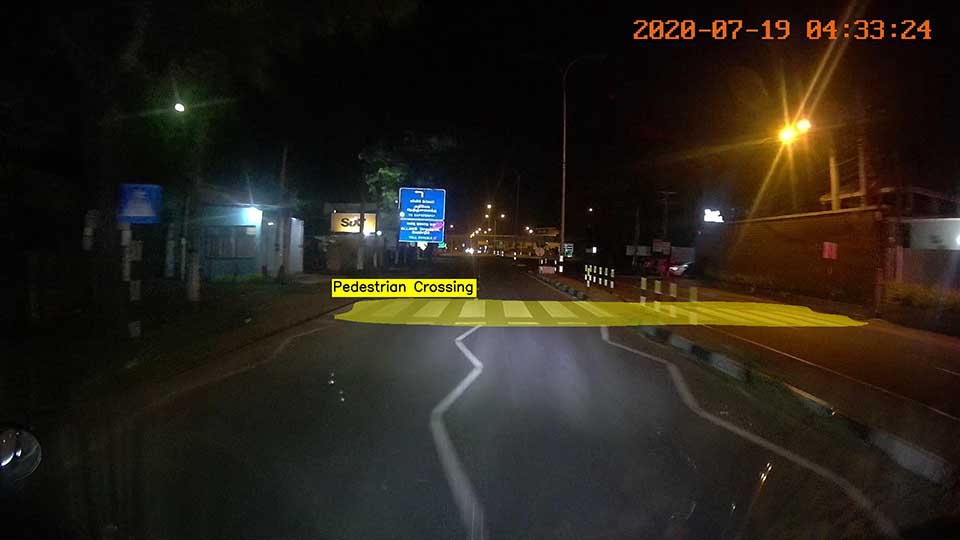}
     \end{subfigure}%
     \begin{subfigure}[b]{0.16\linewidth}
         \centering
         \includegraphics[width=.98\linewidth]{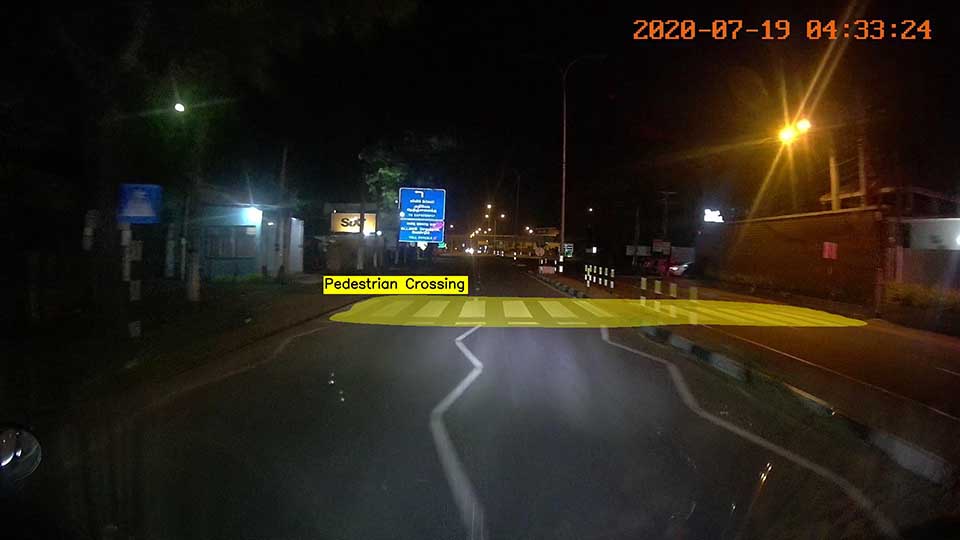}
     \end{subfigure}%
     
     \begin{subfigure}[b]{0.16\linewidth}
         \centering
         \includegraphics[width=.98\linewidth]{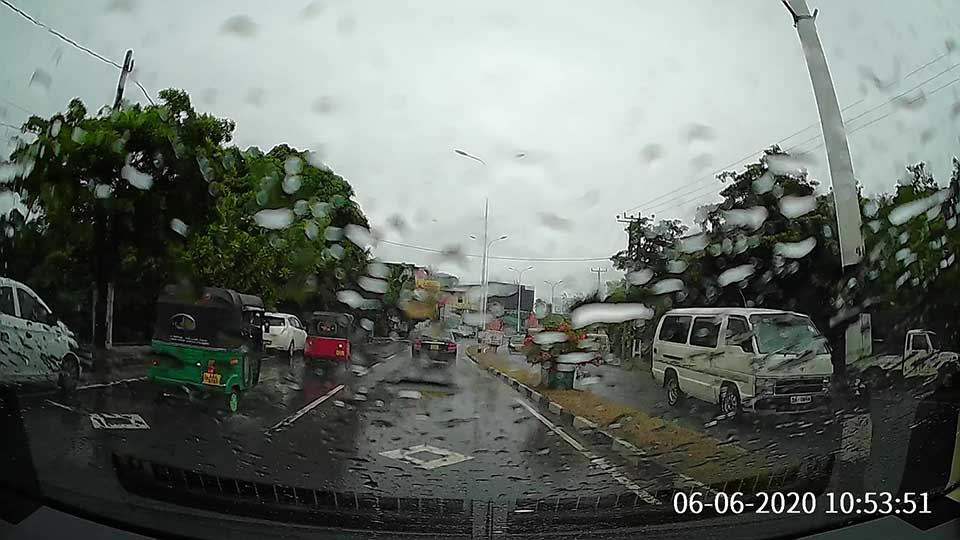}
     \end{subfigure}%
     \begin{subfigure}[b]{0.16\linewidth}
         \centering
         \includegraphics[width=.98\linewidth]{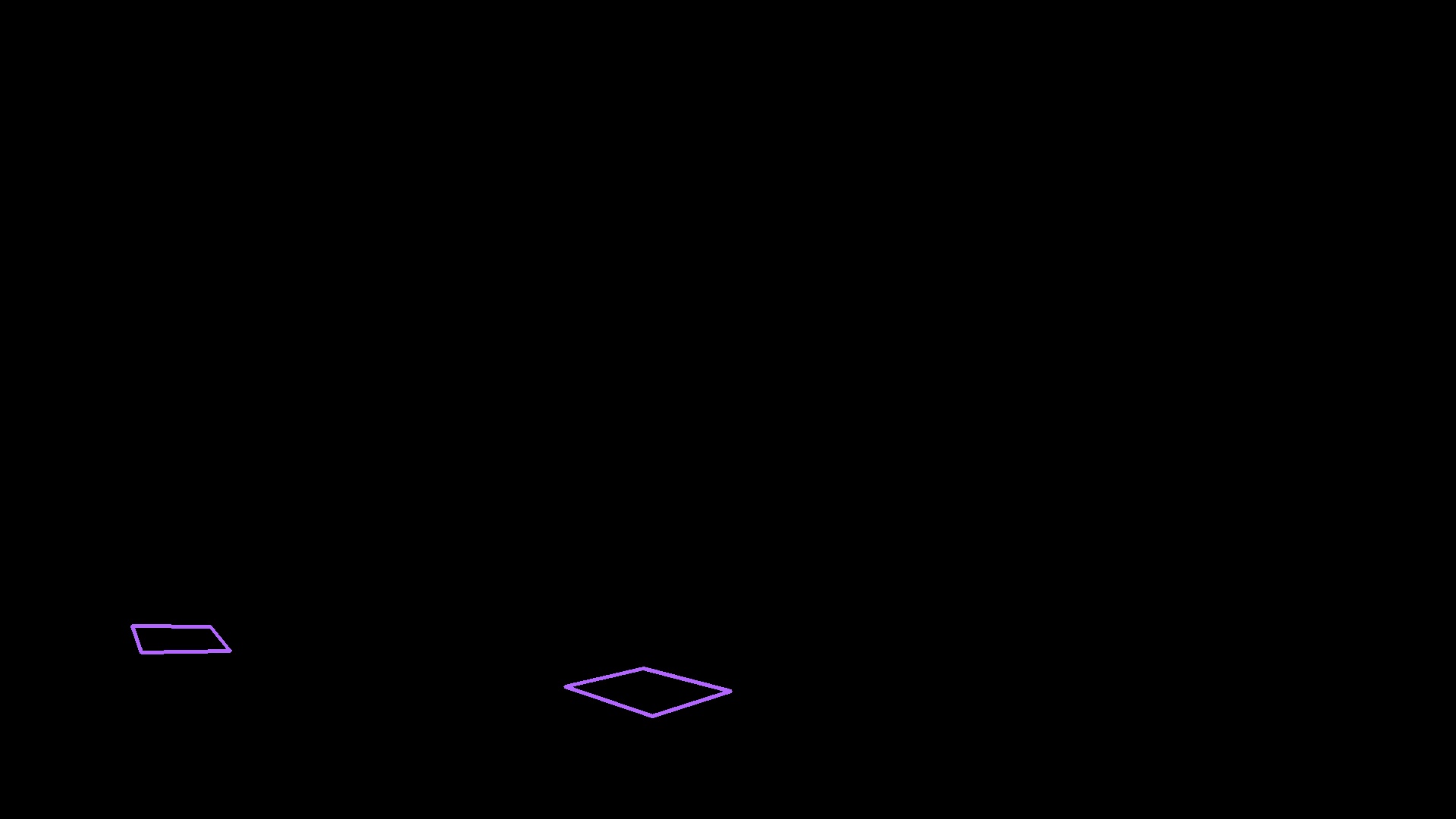}
     \end{subfigure}%
     \begin{subfigure}[b]{0.16\linewidth}
         \centering
         \includegraphics[width=.98\linewidth]{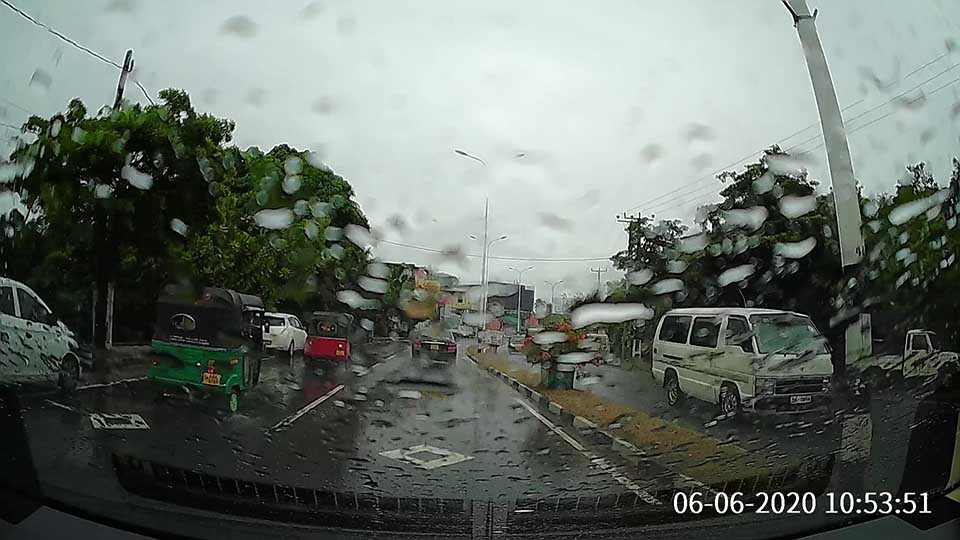}
     \end{subfigure}%
     \begin{subfigure}[b]{0.16\linewidth}
         \centering
         \includegraphics[width=.98\linewidth]{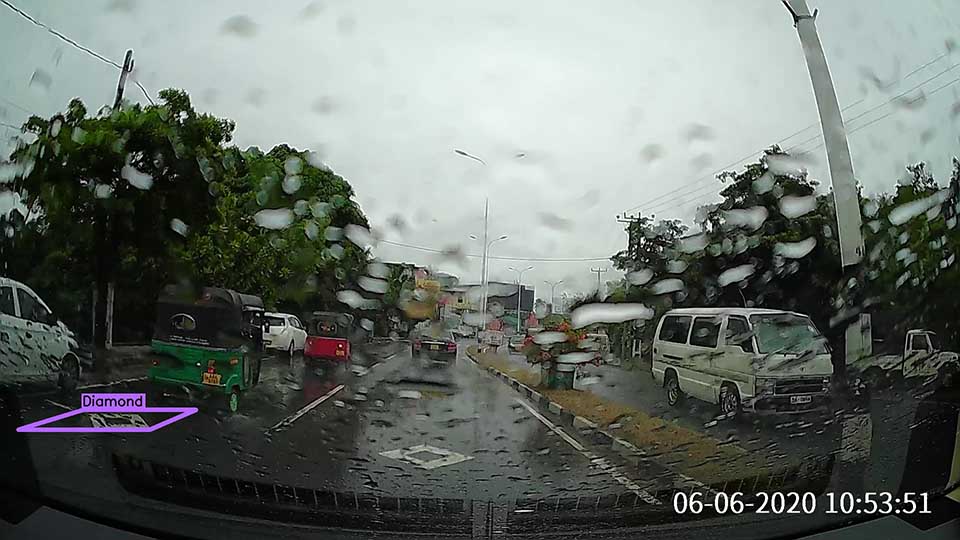}
     \end{subfigure}%
     \begin{subfigure}[b]{0.16\linewidth}
         \centering
         \includegraphics[width=.98\linewidth]{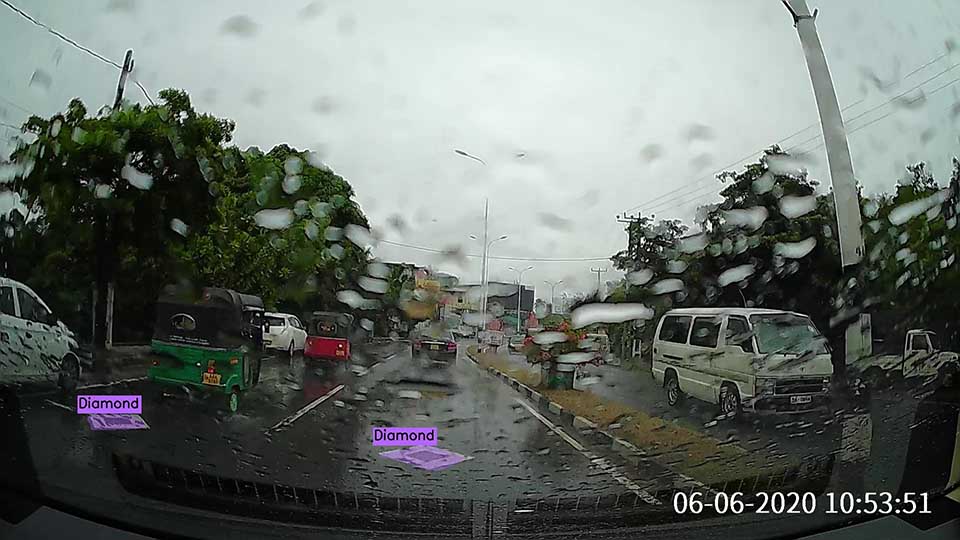}
     \end{subfigure}%
     \begin{subfigure}[b]{0.16\linewidth}
         \centering
         \includegraphics[width=.98\linewidth]{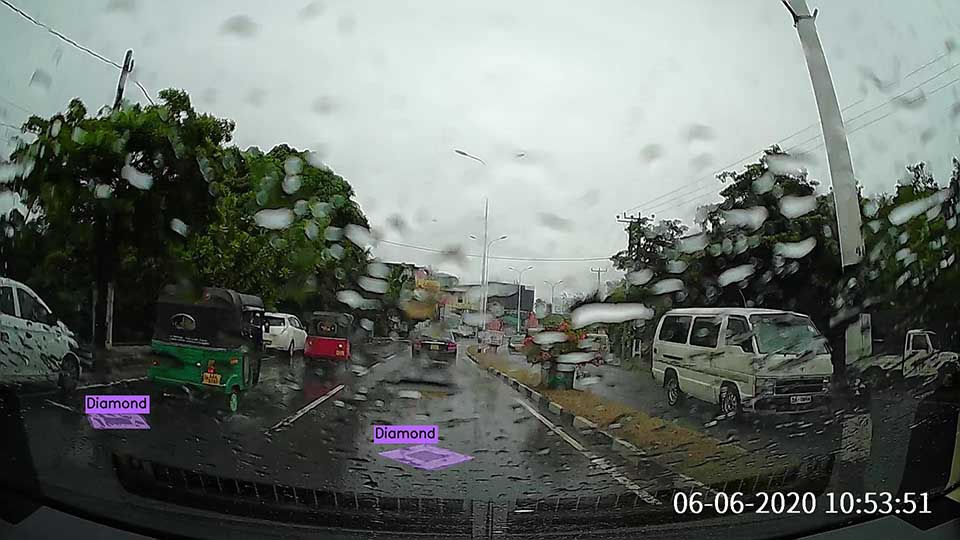}
     \end{subfigure}%
     
     \begin{subfigure}[b]{0.16\linewidth}
         \centering
         \includegraphics[width=.98\linewidth]{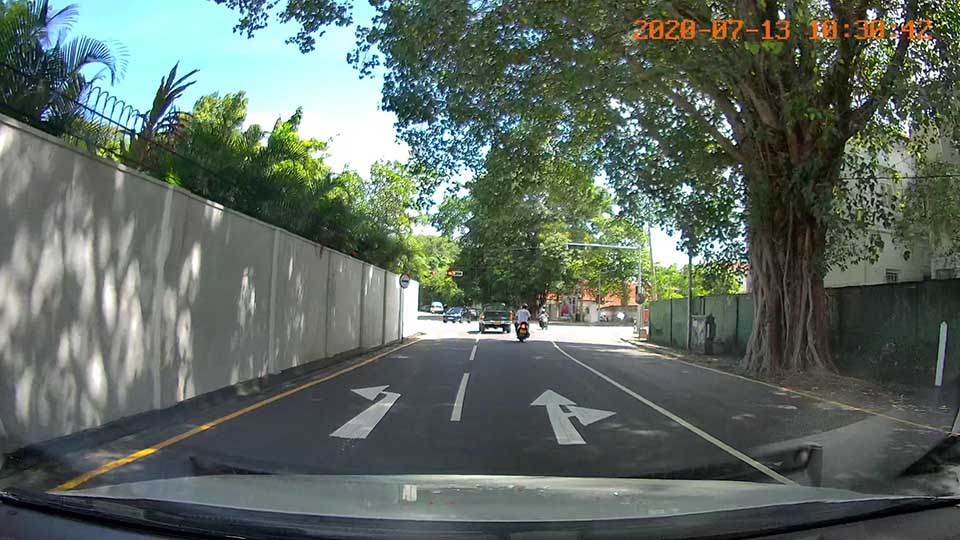}
         \caption{}
     \end{subfigure}%
     \begin{subfigure}[b]{0.16\linewidth}
         \centering
         \includegraphics[width=.98\linewidth]{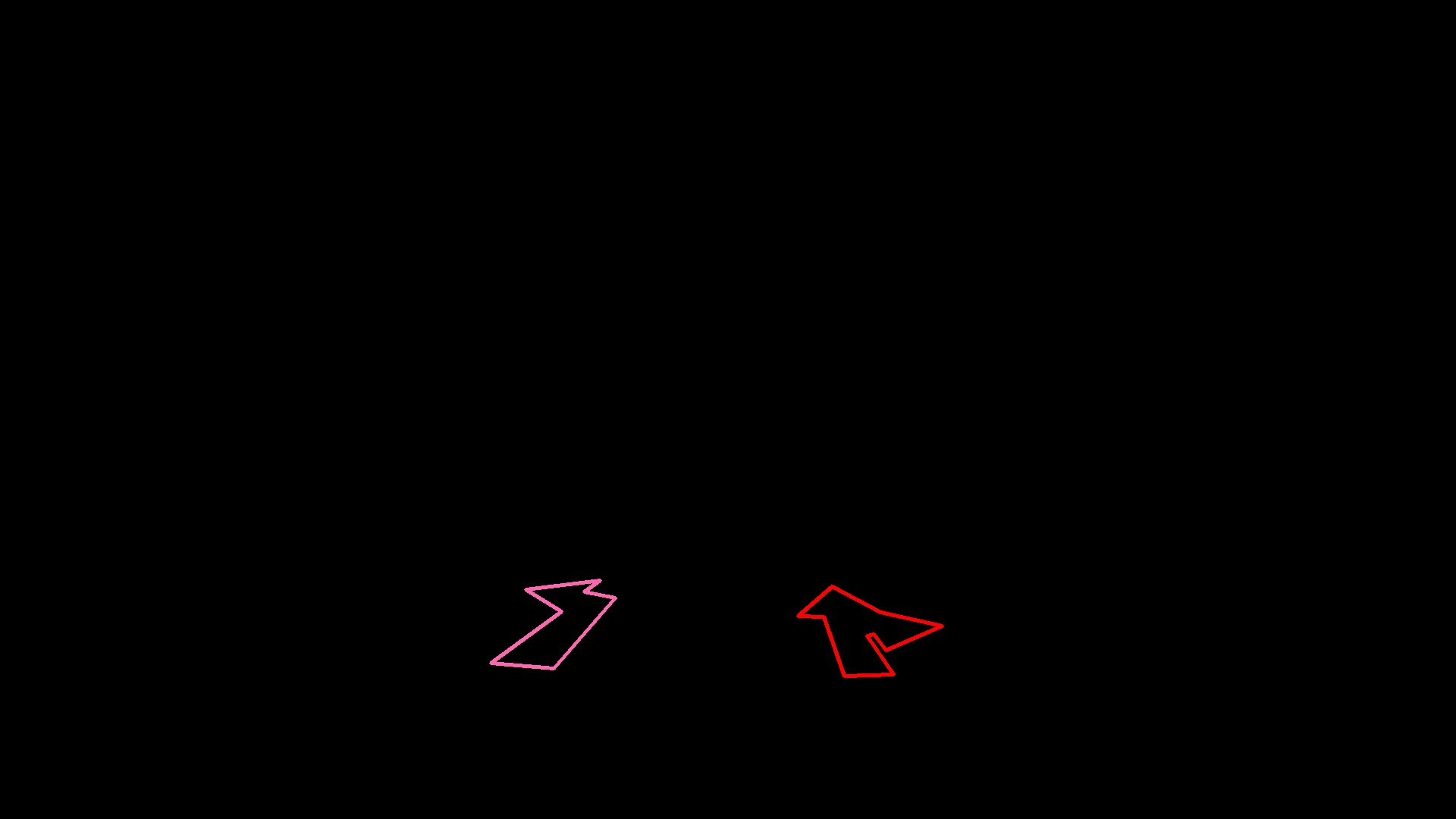}
         \caption{}
     \end{subfigure}%
     \begin{subfigure}[b]{0.16\linewidth}
         \centering
         \includegraphics[width=.98\linewidth]{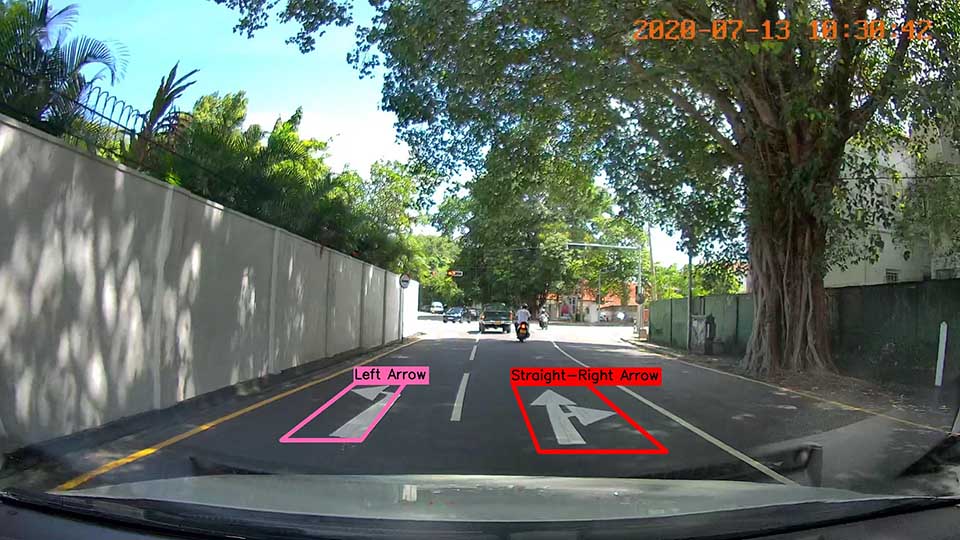}
         \caption{}
     \end{subfigure}%
     \begin{subfigure}[b]{0.16\linewidth}
         \centering
         \includegraphics[width=.98\linewidth]{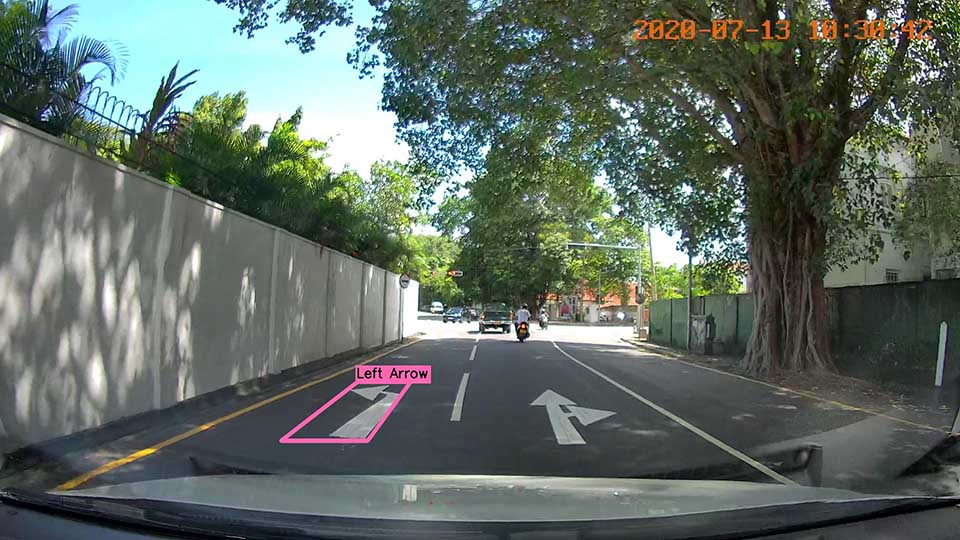}
         \caption{}
     \end{subfigure}%
     \begin{subfigure}[b]{0.16\linewidth}
         \centering
         \includegraphics[width=.98\linewidth]{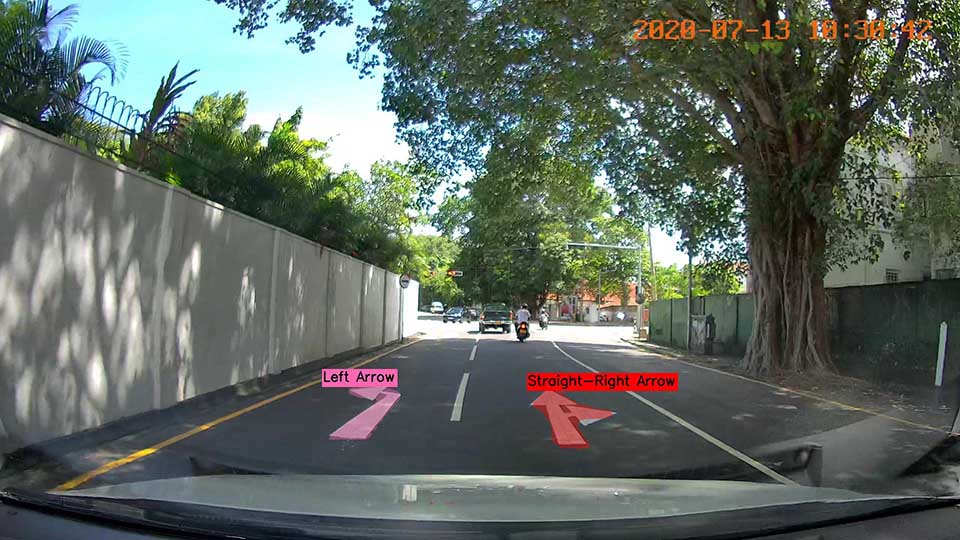}
         \caption{}
     \end{subfigure}%
     \begin{subfigure}[b]{0.16\linewidth}
         \centering
         \includegraphics[width=.98\linewidth]{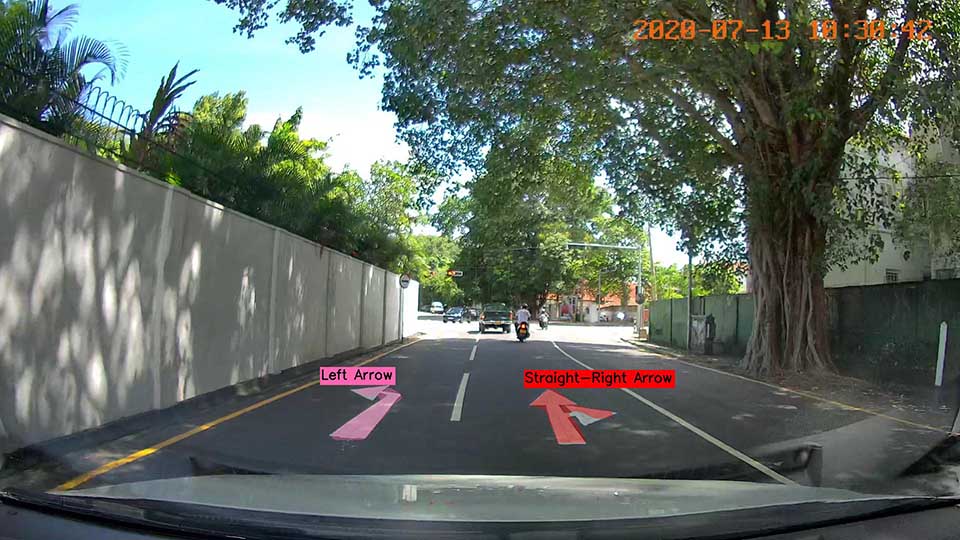}
         \caption{}
     \end{subfigure}%
    \caption{Visualization of road marking detection results on the CeyMo road marking dataset. The top two rows represent detections under normal conditions and the next five rows represent detections under challenging conditions: crowded, dazzle light, night, rain and shadow. (a) Input image (b) Ground truth (c) SSD-MobileNet-v1 \cite{SSD,mobilenet} (d) SSD-Inception-v2 \cite{SSD,inception} (e) Mask-RCNN-Inception-v2 \cite{mask_rcnn,inception} (f) Mask-RCNN-ResNet50 \cite{mask_rcnn,resnet}}
    \label{fi:results}
   \vspace{-0.7em}
\end{figure*}

In this section, we describe the experiments carried out, specifically the data augmentation process and the implementation details.

\subsection{Data Augmentation}

As a step towards mitigating the effect of the class imbalance problem, we follow a simple data augmentation technique during training to increase the number of instances of less frequent signs. It can be observed that left arrow and straight-right arrow classes have comparatively fewer instances than their mirrored classes, right arrow and straight-left arrow. 

Therefore, we horizontally flip the images, which include arrows to obtain the mirrored signs. However, since flipping instances of cycle lane, bus lane and slow road marking classes would lose their meaning, images with those instances are avoided. Furthermore, we randomly change the brightness, saturation, contrast and hue of the input images while training the detection models.




\subsection{Implementation Details}

For the training and testing of our detection algorithms, we use a computational platform comprising an Intel Core i9-9900K CPU and a Nvidia RTX-2080 Ti GPU. We use TensorFlow Object Detection API \cite{huang2017speed} to train the two Mask R-CNN models \cite{mask_rcnn} under the instance segmentation based approach and the two SSD \cite{SSD} models under the object detection based approach.


For the training of the detection models, the following configurations are used. For both SSD-MobileNet-v1 \cite{SSD,mobilenet} and SSD-Inception \cite{SSD,inception} models, RMSProp \cite{ruder2016overview} optimization is used with an initial learning rate of 0.004 and a momentum of 0.9, and the batch size is set to 24. For Mask-RCNN-Inception \cite{mask_rcnn, inception} model, SGD with momentum \cite{pmlr-v28-sutskever13} optimization is used with an initial learning rate of 0.0001 and a momentum of 0.9, and the batch size is set to 4. For Mask-RCNN-ResNet50 \cite{mask_rcnn, resnet} model, SGD with momentum \cite{pmlr-v28-sutskever13} optimization is used with an initial learning rate of 0.0003 and a momentum of 0.9, and the batch size is set to 2.


\section{Results}
\label{sec:results}

In this section, we present the qualitative and quantitative results we obtained. The performance of the two SSD \cite{SSD} based object detection models and the two Mask R-CNN \cite{mask_rcnn} based instance segmentation networks on our CeyMo road marking benchmark dataset is presented in Table \ref{tab:results_scenario} and Table \ref{tab:results_class}. 

Table \ref{tab:results_scenario} shows the $F_{1}$-score values of each model for each of the six categories and the overall $F_{1}$-score for the test set. The inference speed of each model is also listed in frames per second (FPS). It can be observed that Mask R-CNN \cite{mask_rcnn} models under the instance segmentation based approach have been able to achieve better results than the SSD \cite{SSD} models under the object detection based approach. Mask-RCNN-ResNet50 \cite{mask_rcnn,resnet} model has been able to outperform the other models in normal, crowded, dazzle light, rain and shadow categories while Mask-RCNN-Inception-v2 model \cite{mask_rcnn,inception} achieves the highest $F_{1}$-score for the night category. Although the Mask-RCNN-ResNet50 \cite{mask_rcnn,resnet} model has the highest overall $F_{1}$-score of 90.62, its inference speed of 13 FPS is comparatively low which becomes crucial in real-time applications. Mask-RCNN-Inception-v2 \cite{mask_rcnn,inception} model gives a better trade-off between the accuracy and the speed, while SSD-MobileNet-v1 \cite{SSD,mobilenet} and SSD-Inception-v2 \cite{SSD,inception} models along with the inverse perspective transform, result in a moderate accuracy at a higher inference speed. It can be also observed that all models perform better in the normal category and the $F_{1}$-score values are comparatively lower in the five challenging scenarios.

The $F_{1}$-score values for the 11 road marking classes and the Macro $F_{1}$-score of each model are listed in Table \ref{tab:results_class}. Mask-RCNN-ResNet50 \cite{mask_rcnn,resnet} model achieves better results for most of the classes with a Macro $F_{1}$-score of 88.33. It can also be observed that the Macro $F_{1}$-score values of all models are lower than the overall $F_{1}$-score values by around 2 percent or more, which implies that there is a tendency of the models to perform better in certain classes than others. Pedestrian crossings which can be found frequently within the dataset and capture a larger area of the road, are well detected by all models. Classes like slow, bus lane and cycle lane have comparatively lower number of instances in the dataset. Nevertheless, all four models have been able to detect those classes with a good accuracy, due to the distinct shapes and features of those classes. Although we increase the number of arrow signs in the train set through the data augmentation process, the accuracy values of arrow signs are low, when compared with other road markings. This can be mostly due to the similarity of arrow sign classes within themselves, as well as with the lane markings on the road surface. 

Qualitative results obtained by our two object detection models and the two instance segmentation networks are visualized in Figure \ref{fi:results}, along with the input images and the ground truth for the six categories in the test set. It can be observed that the Mask R-CNN \cite{mask_rcnn} models under the instance segmentation based approach perform better, especially in challenging scenarios. Furthermore, the segmentation masks used in those models result in more precise localization of road markings than 4-sided polygons used in the object detection based approach.




\section{Conclusion}
\label{sec:conclusion}

In this work, we introduced the CeyMo road marking dataset for road marking detection, addressing the limitations present in the existing datasets. The novel benchmark dataset consists of 2887 images taken under different traffic, lighting and weather conditions, covering 4706 road marking instances belonging to 11 road marking classes. We provide road marking annotations as polygons, bounding boxes and segmentation masks to facilitate a wide range of road marking detection algorithms. The evaluation metrics provided along with the evaluation script will enable direct comparison of future work done on road marking detection. Furthermore, we evaluated the effectiveness of road marking detection firstly, using object detectors on inverse perspective transformed images and secondly, using end-to-end instance segmentation based networks. The speed and accuracy scores for two object detectors and two instance segmentation network architectures are provided as a performance baseline for the benchmark dataset. We believe that the CeyMo road marking dataset can be used to design and evaluate novel road marking detection algorithms stepping towards real-time, accurate road marking detection in challenging environments in the future.    

{\small
\bibliographystyle{ieee_fullname}
\bibliography{egbib}

\begin{thebibliography}{10}\itemsep=-1pt

\bibitem{wacv}
O. {Bailo}, S. {Lee}, F. {Rameau}, J.~S. {Yoon}, and I.~S. {Kweon}.
\newblock Robust road marking detection and recognition using density-based
  grouping and machine learning techniques.
\newblock In {\em IEEE Winter Conference on Applications of Computer Vision
  (WACV)}, pages 760--768, 2017.

\bibitem{chan2015pcanet}
T. {Chan}, K. {Jia}, S. {Gao}, J. {Lu}, Z. {Zeng}, and Y. {Ma}.
\newblock Pcanet: A simple deep learning baseline for image classification?
\newblock {\em IEEE Transactions on Image Processing}, 24(12):5017--5032, 2015.

\bibitem{chen2015road}
T. {Chen}, Z. {Chen}, Q. {Shi}, and X. {Huang}.
\newblock Road marking detection and classification using machine learning
  algorithms.
\newblock In {\em IEEE Intelligent Vehicles Symposium (IV)}, pages 617--621,
  2015.

\bibitem{cheng2014bing}
M. {Cheng}, Z. {Zhang}, W. {Lin}, and P. {Torr}.
\newblock Bing: Binarized normed gradients for objectness estimation at 300fps.
\newblock In {\em IEEE Conference on Computer Vision and Pattern Recognition},
  pages 3286--3293, 2014.

\bibitem{Cordts2016Cityscapes}
M. {Cordts}, M. {Omran}, S. {Ramos}, T. {Rehfeld}, M. {Enzweiler}, R.
  {Benenson}, U. {Franke}, S. {Roth}, and B. {Schiele}.
\newblock The cityscapes dataset for semantic urban scene understanding.
\newblock In {\em IEEE Conference on Computer Vision and Pattern Recognition
  (CVPR)}, pages 3213--3223, 2016.

\bibitem{ding2020comprehensive}
L. {Ding}, H. {Zhang}, J. {Xiao}, B. {Li}, S. {Lu}, R. {Klette}, M.
  {Norouzifard}, and F. {Xu}.
\newblock A comprehensive approach for road marking detection and recognition.
\newblock {\em Multimedia Tools and Applications}, 79(23):17193--17210, 2020.

\bibitem{Greenhalgh}
J. {Greenhalgh} and M. {Mirmehdi}.
\newblock Automatic detection and recognition of symbols and text on the road
  surface.
\newblock In {\em International Conference on Pattern Recognition and Methods
  (ICPRM)}, pages 124--140, 2015.

\bibitem{mask_rcnn}
K. {He}, G. {Gkioxari}, P. {Dollár}, and R. {Girshick}.
\newblock Mask r-cnn.
\newblock In {\em IEEE International Conference on Computer Vision (ICCV)},
  pages 2980--2988, 2017.

\bibitem{resnet}
K. {He}, X. {Zhang}, S. {Ren}, and J. {Sun}.
\newblock Deep residual learning for image recognition.
\newblock In {\em IEEE Conference on Computer Vision and Pattern Recognition
  (CVPR)}, pages 770--778, 2016.

\bibitem{mobilenet}
A.G. {Howard}, M. {Zhu}, B. {Chen}, D. {Kalenichenko}, W. {Wang}, T. {Weyand},
  M. {Andreetto}, and H. {Adam}.
\newblock Mobilenets: Efficient convolutional neural networks for mobile vision
  applications.
\newblock {\em arXiv preprint arXiv:1704.04861}, 2017.

\bibitem{huang2017speed}
J. {Huang}, V. {Rathod}, C. {Sun}, M. {Zhu}, A. {Korattikara}, A. {Fathi}, I.
  {Fischer}, Z. {Wojna}, Y. {Song}, S. {Guadarrama}, et~al.
\newblock Speed/accuracy trade-offs for modern convolutional object detectors.
\newblock In {\em IEEE Conference on Computer Vision and Pattern Recognition
  (CVPR)}, pages 7310--7311, 2017.

\bibitem{Kheyrollahi}
A. {Kheyrollahi} and T. {Breckon}.
\newblock Automatic real-time road marking recognition using a feature driven
  approach.
\newblock {\em Machine Vision and Applications}, 23:123--133, 2010.

\bibitem{VPGNet}
S. {Lee}, J. {Kim}, J.~S. {Yoon}, S. {Shin}, O. {Bailo}, N. {Kim}, T. {Lee},
  H.~S. {Hong}, S. {Han}, and I.~S. {Kweon}.
\newblock Vpgnet: Vanishing point guided network for lane and road marking
  detection and recognition.
\newblock In {\em IEEE International Conference on Computer Vision (ICCV)},
  pages 1965--1973, 2017.

\bibitem{IPMbased12}
H. {Li}, M. {Feng}, and X. {Wang}.
\newblock Inverse perspective mapping based urban road markings detection.
\newblock {\em IEEE 2nd International Conference on Cloud Computing and
  Intelligence Systems}, 03:1178--1182, 2012.

\bibitem{SSD}
W. {Liu}, D. {Anguelov}, D. {Erhan}, C. {Szegedy}, S. {Reed}, C.Y. {Fu}, and A.
  {Berg}.
\newblock Ssd: Single shot multibox detector.
\newblock In {\em European Conference on Computer Vision (ECCV)}, pages 21--37,
  2016.

\bibitem{TROM}
X. {Liu}, Z. {Deng}, H. {Lu}, and L. {Cao}.
\newblock Benchmark for road marking detection: Dataset specification and
  performance baseline.
\newblock In {\em IEEE 20th International Conference on Intelligent
  Transportation Systems (ITSC)}, pages 1--6, 2017.

\bibitem{MSER}
J. {Matas}, O. {Chum}, M. {Urban}, and T. {Pajdla}.
\newblock Robust wide baseline stereo from maximally stable extremal regions.
\newblock {\em Image and Vision Computing}, 22:761--767, 2004.

\bibitem{review}
A. {Morrissett} and S. {Abdelwahed}.
\newblock A review of non-lane road marking detection and recognition.
\newblock In {\em IEEE 23rd International Conference on Intelligent
  Transportation Systems (ITSC)}, pages 1--7, 2020.

\bibitem{FRCNN}
S. {Ren}, K. {He}, R. {Girshick}, and J. {Sun}.
\newblock Faster r-cnn: Towards real-time object detection with region proposal
  networks.
\newblock {\em IEEE Transactions on Pattern Analysis and Machine Intelligence},
  39(6):1137--1149, 2017.

\bibitem{ruder2016overview}
S. {Ruder}.
\newblock An overview of gradient descent optimization algorithms.
\newblock {\em arXiv preprint arXiv:1609.04747}, 2016.

\bibitem{Srikanthan}
S. {Suchitra}, R.~K. {Satzoda}, and T. {Srikanthan}.
\newblock Detection classification of arrow markings on roads using signed edge
  signatures.
\newblock In {\em IEEE Intelligent Vehicles Symposium (IV)}, pages 796--801,
  2012.

\bibitem{Suhr2015FastSR}
J.~K. {Suhr} and H. {Jung}.
\newblock Fast symbolic road marking and stop-line detection for vehicle
  localization.
\newblock In {\em IEEE Intelligent Vehicles Symposium (IV)}, pages 186--191,
  2015.

\bibitem{pmlr-v28-sutskever13}
I. {Sutskever}, J. {Martens}, G. {Dahl}, and G. {Hinton}.
\newblock On the importance of initialization and momentum in deep learning.
\newblock In {\em 30th International Conference on Machine Learning (ICML)},
  pages 1139--1147, 2013.

\bibitem{inception}
C. {Szegedy}, V. {Vanhoucke}, S. {Ioffe}, J. {Shlens}, and Z. {Wojna}.
\newblock Rethinking the inception architecture for computer vision.
\newblock In {\em IEEE Conference on Computer Vision and Pattern Recognition
  (CVPR)}, pages 2818--2826, 2016.

\bibitem{labelme2016}
K. {Wada}.
\newblock {labelme: Image Polygonal Annotation with Python}.
\newblock \url{https://github.com/wkentaro/labelme}, 2016.

\bibitem{ananth}
T. {Wu} and A. {Ranganathan}.
\newblock A practical system for road marking detection and recognition.
\newblock In {\em IEEE Intelligent Vehicles Symposium (IV)}, pages 25--30,
  2012.

\bibitem{BDD}
F. {Yu}, H. {Chen}, X. {Wang}, W. {Xian}, Y. {Chen}, F. {Liu}, V. {Madhavan},
  and T. {Darrell}.
\newblock Bdd100k: A diverse driving dataset for heterogeneous multitask
  learning.
\newblock In {\em IEEE/CVF Conference on Computer Vision and Pattern
  Recognition (CVPR)}, pages 2633--2642, 2020.

\bibitem{Zhe_2016_CVPR}
Z. {Zhu}, D. {Liang}, S. {Zhang}, X. {Huang}, B. {Li}, and S. {Hu}.
\newblock Traffic-sign detection and classification in the wild.
\newblock In {\em IEEE Conference on Computer Vision and Pattern Recognition
  (CVPR)}, pages 2110--2118, 2016.

\end{thebibliography}
}

\end{document}